\pdfoutput=1

\documentclass[11pt,table]{article}

\usepackage[]{acl}

\usepackage{times}
\usepackage{makecell}
\usepackage{latexsym}
\usepackage{graphicx}
\usepackage{soul}
\usepackage{multirow}
\usepackage{booktabs}
\usepackage{tabularx}
\usepackage{arydshln}
\usepackage{array}
\usepackage{hhline}
\usepackage{float}
\usepackage{makecell}
\usepackage{arydshln}
\usepackage{siunitx} 
\usepackage{tikz}
\usepackage{amsmath}
\usepackage{cleveref}
\usepackage{subcaption}
\usepackage{lipsum}  
\usepackage{xfrac}
\usepackage[colorinlistoftodos]{todonotes}
\usepackage{duckuments}
\definecolor{lightgray}{gray}{0.95}
\usepackage{nicefrac}

\newcommand{\wideunderline}[2][2em]{%
  \underline{\makebox[\ifdim\width>#1\width\else#1\fi]{#2}}%
}

\makeatletter
\g@addto@macro{\endtabular}{\gdef\rowfonttype{}}
\makeatother
\newcommand{\rowfonttype}{}
\newcolumntype{L}{>{\rowfonttype\strut}l}
\crefformat{section}{\S#2#1#3} 
\crefformat{subsection}{\S#2#1#3}
\crefformat{subsubsection}{\S#2#1#3}
\usepackage[T1]{fontenc}
\usepackage{booktabs} 
\usepackage{caption}

\usepackage[utf8]{inputenc}

\usepackage{microtype}

\newcommand{\mf}{{MFSC }}

\newboolean{isRed}
\setboolean{isRed}{True} 
\newcommand{\rd}[1]{%
    \ifthenelse{\boolean{isRed}}%
    {\textcolor{red}{#1}}
    {#1}
}

\newboolean{isoldbb}
\setboolean{isoldbb}{False} 
\newcommand{\blue}[1]{%
    \ifthenelse{\boolean{isoldbb}}%
    {\textcolor{blue}{#1}}
    {#1}
}
\newboolean{isbb}
\setboolean{isbb}{False} 
\newcommand{\newblue}[1]{%
    \ifthenelse{\boolean{isbb}}%
    {\textcolor{blue}{#1}}
    {#1}
}

\newboolean{emnlp_bool}
\setboolean{emnlp_bool}{False} 
\newcommand{\emnlp}[1]{%
    \ifthenelse{\boolean{emnlp_bool}}%
    {\textcolor{blue}{#1}}
    {#1}
}

\usepackage[most]{tcolorbox} 
\title{Cost-Efficient Subjective Task Annotation and Modeling \\  through Few-Shot Annotator Adaptation}

\author{Preni Golazizian \And Alireza S. Ziabari \And Ali Omrani \And Morteza Dehghani \AND \normalfont \small
University of Southern California \\
\texttt {\small\{golazizi, salkhord, aomrani, mdehghan\}@usc.edu}
}

\begin{document}
\maketitle

\begin{abstract}

In subjective NLP tasks, where a single ground truth does not exist, the inclusion of diverse annotators becomes crucial as their unique perspectives significantly influence the annotations. In realistic scenarios, the annotation budget often becomes the main determinant of the number of perspectives (i.e., annotators) included in the data and subsequent modeling. We introduce a novel framework for annotation collection and modeling in subjective tasks that aims to minimize the annotation budget while maximizing the predictive performance for each annotator. Our framework has a two-stage design: first, we rely on a small set of annotators to build a multitask model, and second, we augment the model for a new perspective by strategically annotating a few samples per annotator. To test our framework at scale, we introduce and release a unique dataset, Moral Foundations Subjective Corpus, of 2000 Reddit posts annotated by 24 annotators for moral sentiment. 
We demonstrate that our framework surpasses the previous SOTA in capturing the annotators' individual perspectives with as little as 25\% of the original annotation budget on two datasets. Furthermore,  our framework results in more equitable models, reducing the performance disparity among annotators.


\end{abstract}

\section{Introduction}

The common pipeline for supervised learning in Natural Language Processing (NLP) starts by collecting annotations from multiple annotators. These annotations are often aggregated through majority voting \citep{waseem2016hateful} to construct a \textit{ground truth} or \textit{gold standard} on which the subsequent modeling is performed.  
In recent years, researchers have advocated for a transition from single ground-truth labels to annotator-level modeling, aiming to capture diverse perspectives, enhance contextual understanding, and incorporate cultural nuances \citep{uma2021learning}, and have proposed different frameworks that take into account unique perspectives of the annotators by modeling them as separate subtasks \citep{davani2022dealing, kanclerz2022if}.



The impact of individual annotators' backgrounds and life experiences on annotations in subjective tasks signifies the importance of incorporating a diverse set of annotators. Nevertheless, the primary constraint on achieving this diversity is often the annotation budget, limiting the number and, consequently, the diversity of perspectives considered. 
In this paper,  we introduce a novel framework for annotation collection and modeling in subjective tasks. Our framework is designed to minimize the annotation budget required to model a fixed number of annotators, while maximizing the predictive performance for each annotator. 



Our framework operates in two stages. In the first stage, data is collected from a small pool of annotators. This data serves as a foundation for building a multitask model that captures the general patterns for the task and provides a signal of differences among individual annotators. Informed by the first stage annotations, the second stage involves collecting a few samples from each new annotator that best capture their differences from the general patterns. We use this data to augment the model from the first stage to learn the new annotators' perspective from a few examples (\autoref{fig:main}).

We introduce a unique dataset that enables the study of detecting moral content, an understudied subjective task, at a scale that was not possible before\footnote{The dataset will be released as part of the accepted paper}. The Moral Foundations Subjective Corpus (MFSC) is a collection of 2000 Reddit posts, each annotated by 24 annotators for moral content along with annotators' responses to a range of psychological questionnaires (\S \ref{sec:data}).We use the MFSC dataset in conjunction with the Brexit Hate Dataset \citep{akhtar2021whose} to extensively study each component of our proposed framework.
We evaluate our framework on three models: RoBERTa-Base, RoBERTa-Large, and Llama-3.
In section \ref{sec:res_main}, we demonstrate the efficacy of our framework in capturing diverse annotator perspectives under budget constraints. Our framework achieves a 4\% increase in $F_1$ score with access to just 50\% of the annotation budget in hate speech detection, and a 2\% gain in moral sentiment detection with as little as 25\% of the original annotation budget. 
Furthermore, we evaluate the efficiency of our framework in scaling to more annotators, i.e., incorporating a new annotator into an already existing annotated dataset and model through our second-stage few-shot adaptation. Our results show an $F_1$ score gain of 9\% and 4\% in the Brexit and MFSC datasets, respectively, demonstrating the scalability of our framework. 
Next, in section \ref{sec:dispar}, we show that our proposed framework yields a more equitable model by minimizing performance disparity across annotators. Specifically, in the lowest budget scenarios, our approach reduces the standard deviation of the performance across annotators by 7\% in hate speech detection and by 1\% in moral sentiment classification. Finally, in section \ref{sec:ann_level}, we extend our analysis to investigate whether the selection of the initial set of annotators in the first stage of our framework affects the model's performance.



Our experiments on two subjective datasets revealed that our framework consistently surpasses previous state-of-the-art models with access to as little as 25\% of the original annotation budget. In addition, our framework produced more equitable models with reduced performance disparities among the annotators.
By minimizing data requirements, our cost-efficient framework for subjective tasks enables us to scale the number of included annotators and, hence, improve the diversity of captured perspectives. Furthermore, the two-stage design of our framework facilitates the integration of new annotators into pre-existing datasets.

\section{Related Work}

\noindent\textbf{Subjective Tasks in NLP:}
In recent years, the variety of tasks for which NLP is used has significantly expanded. In many of these tasks,  a single ground truth does not exist, making them inherently \textit{subjective} in nature. In subjective tasks, researchers have argued that disagreements in particular labels should not be treated as statistical noise \citep{larimore2021reconsidering, pavlick2019inherent, plank-2022-problem}, as they are often indicative of individual differences which are driven by different backgrounds and lived experiences of the annotators \citep{akhtar2019new, plank2014linguistically, prabhakaran2021releasing, diaz2018addressing, garten2019incorporating, ferracane-etal-2021-answer}.
For example, \citet{davani2023hate} revealed how the stereotypes of annotators influence their behavior when annotating hate speech. In a similar context, \citet{sap2021annotators} demonstrate that annotators' identity and beliefs impact their ratings of toxicity. \citet{sang2022origin} conducted a study showing that differences in age and personality among annotators result in variations in their annotations. \citet{larimore2021reconsidering} explored how annotators' perceptions of racism differ based on their own racial identity.
\citet{basile2020s} calls for a paradigm shift away from majority aggregated ground truths, and towards representative frameworks preserving unique perspectives of the annotators. In their later work, \citet{basile2021toward} define the phenomena of \emph{Data Perspectivism}, and share recommendations and outlines to advance the perspectivist stance in machine learning.

\noindent\textbf{Capturing the Perspectives:}
One method for learning directly from crowd annotations is using soft loss, where the probability distributions of item labels are used as soft targets in a loss function \citep{peterson2019human}. However, this approach does not provide individual predictions for annotators, making it unsuitable for subjective tasks that require such specificity.
To capture annotator-level labels, \citet{akhtar2020modeling} proposed dividing annotators into groups based on similar personal characteristics and creating different sets of gold standards for each group. \citet{kanclerz2022if} and \citet{deng-etal-2023-annotate} incorporated knowledge about annotators into their models to make them personalized. \citet{davani2022dealing} propose a multitask approach, modeling each annotators' perspective as a subtask, while having a shared encoder across the subtasks. 
\citet{baumler2023examples} and  \citet{wang2023actor}  propose active learning methods for reducing the budget of data collection by proposing methods for collecting samples based on model confidence and annotators' disagreement. \citet{casola2023confidence} also proposes ensembling perspective-aware models based on their confidence.
 %
\section{Method}
\label{sec:method}
\begin{figure*}[t!]
    \centering
    \includegraphics[width=\linewidth]{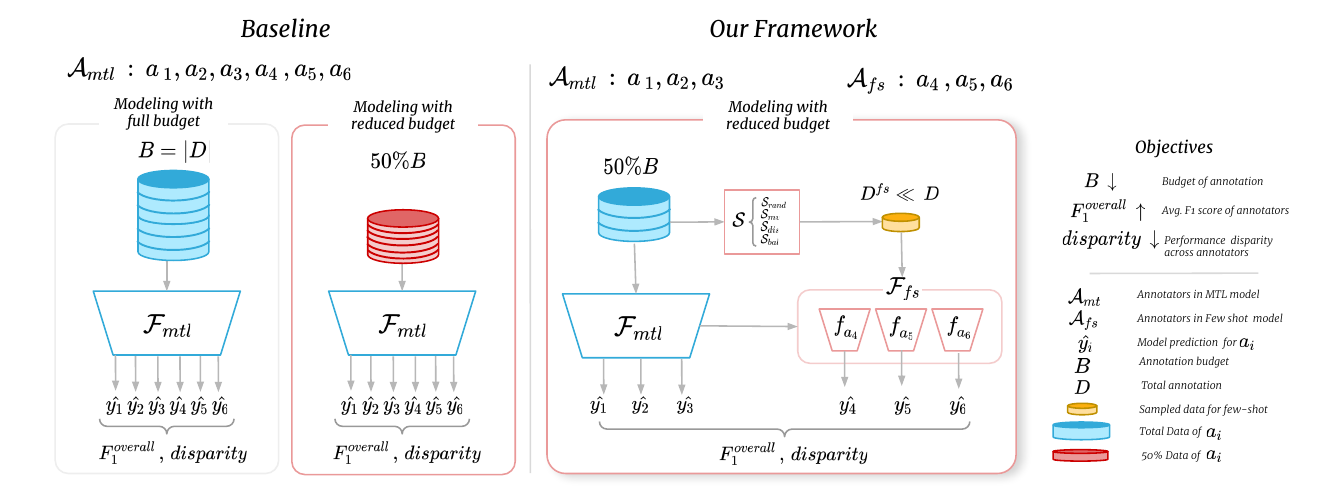}
    \caption{\textbf{Left:} The baseline approach for annotator-level modeling, in full and reduced budget scenarios. \textbf{Right:} Our two-stage proposed framework, designed to achieve the outlined objectives}
    \label{fig:main}
  
\end{figure*}
\noindent\textbf{Problem Formulation:}
To formalize the task, suppose we have a set of annotators $\mathcal{A} = \{a_1, ... , a_n\}$ and input texts $X = \{x_1, x_2, ... , x_m\}$ and their corresponding annotations $Y = \{y_1, y_2, ... , y_m\}$. Let $D=\{D_{a_i} | a_i \in \mathcal{A}\}$ be the entire annotations and $D_{a_i} = \{X_{a_i}, Y_{a_i}\}$ denote data collected from annotator $a_i$. Then the budget $B=|D|$ is defined as the total number of annotations collected. Let $\mathcal{F}=\{f_{a_i}) | a_i \in \mathcal{A} \}$
and $f_{a_i}$ denote the model capturing labels assigned by annotator $a_i$. \\
\noindent\textbf{Proposed Framework:}
We design our framework with two objectives: first, maximizing the average performance over all annotators. Second, minimizing the budget ($B$) required to achieve the first goal. The second objective allows us to increase the number of annotators' perspectives ($|\mathcal{A}|$ ) captured with a given budget.
Our framework design is based on two key intuitions. Firstly, as shown in \autoref{fig:all_results}, multitask learning (the orange line), which has often been treated as the upper bound by previous work, does not always improve in performance as the number of annotators grows. Secondly, even in subjective tasks, there exists a substantial number of texts on which annotators mostly agree, particularly when these texts are randomly drawn from a source. Therefore, obtaining many annotations on such instances is not beneficial in learning a new perspective. In line with these intuitions, our framework consists of two stages 
(\autoref{fig:main}). In the first stage, we learn the commonalities between annotators through a multitask model $\mathcal{F}_{mtl}$. A crucial difference of our approach in comparison to previous multitask methods is that we only collect annotations from a small subset of annotators $\mathcal{A}_{mtl} \subset \mathcal{A}$. In the second stage, we learn the perspectives of new annotators $\mathcal{A}_{fs} = \mathcal{A} - \mathcal{A}_{mtl}$ with only a few shots. Specifically, we collect annotations for $k$ input texts $\mathcal{S}(X) \subset X$, where $\mathcal{S}$ is a sampling function that ideally helps in capturing patterns specific to individual annotators' perspectives. Let $D^{fs}_{a_i} = \{(x,y_{ai}) | x \in \mathcal{S}(X)\}$ and $|D^{fs}_{a_i}| = k << |D_{a_i}|$. We initialize $\mathcal{F}_{\mathcal{A}_{fs}}$ with $\mathcal{F}_{mtl}$ and train it on $D^{fs}_{a_i}$. 

\noindent\textbf{Sampling Function ($\mathcal{S}$)}: We explore four different sampling functions: 1) $\mathcal{S}_{rand}$: selects a random sample for each annotator 2) $\mathcal{S}_{mv}$: 
selects a balanced sample determined by the majority vote of the annotators. For a set of annotators $\mathcal{A}_{mtl}$, we calculate the majority vote among these annotators and select $k$ samples that have an equal number of each label based on that majority vote.
3) $\mathcal{S}_{dis}$ selects the samples from $\mathcal{A}_{mtl}$ with highest disagreement score, and  
 4) $\mathcal{S}_{bal}$ acts as an oracle, selecting a balanced sample based on a specific annotator's label, not the majority vote. Therefore, if we have a new annotator, $\mathcal{S}_{bal}$ would select a balanced sample based on the annotations of that specific annotator. One frequent challenge in some subjective tasks is the heavy imbalance in class frequencies. Hence, we chose $\mathcal{S}_{mv}$ and $\mathcal{S}_{bal}$ to provide a more balanced sample to the few-shot model for each annotator. We added $\mathcal{S}_{dis}$ with the goal of providing samples that differentiate the individual annotator perspectives to the model. We use the  ``item disagreement'' and ``annotator disagreement'' measures from  \citet{davani2023hate} to select samples in $\mathcal{S}_{dis}$.

\section{Experiments}

\subsection{Datasets}
\label{sec:data}
\blue{We run experiments on two datasets annotated for subjective tasks: Brexit Hate dataset \citep{akhtar2021whose} and the Moral Foundations Subjective Corpus (MFSC), which we created as part of this work to explore this less-studied subjective task. Both datasets contain per-annotator labels for instances, with every instance being annotated by all annotators. This ensures that any observed performance gains are attributed to our method, rather than the specific samples annotated by each annotator.
Additionally, we evaluate our framework on the Gab Hate Corpus \citep[GHC;][]{kennedy2018gab}, where the number of annotations by different annotators varies. Detailed experiments and results for this dataset are presented in Appendix \ref{sec:ghc}.}


\noindent\textbf{Moral Foundations Subjective Corpus (MFSC)}: 
\noindent We introduce the Moral Foundations Subjective Corpus (MFSC), a new dataset consisting of 2000 Reddit posts annotated by 24 annotators for moral sentiment based on the Moral Foundations Theory \citep[MFT;][]{graham2013moral, atari2023morality}. Morality, being a subjective concept heavily influenced by cultural backgrounds \cite{graham2016cultural}, has not been extensively explored in the NLP community. 

Each sample in the MFSC is annotated with a binary label indicating moral sentiment: \textit{1} if the sentence pertains to morality and \textit{0} if it does not. We utilize this binary moral/non-moral label in our experiments. Additionally, we have collected more fine-grained labels of morality, which are detailed in the Appendix \ref{sec:app-data-stats}. 
Examples of the dataset and their annotations for moral sentiment are presented in \autoref{tab:example_annotations}. The demographics of the annotators are provided in Appendix \ref{sec:mfrc-diverse}.

\renewcommand{\arraystretch}{1.2}
\begin{table}[ht]
\centering
\resizebox{\columnwidth}{!}{
\begin{tabular}{@{}p{8.5cm} c c c c c c@{}}
\toprule
\textbf{MFSC examples} & \textbf{a$_1$} & \textbf{a$_2$} & \textbf{a$_3$} & \textbf{a$_4$} & \textbf{a$_5$} & \textbf{a$_6$} \\
\midrule
You're a horrible person, and deserve the same thing to happen to you. & 1 & 1 & 1 & 1 & 1 & 1 \\
As an expat Brit, I was moved: What a brilliant unifying speech. Here's fingers crossed for you USA. & 1 & 0 & 0 & 1 & 1 & 1 \\
That meal is insane compared to what we got. Don't think we ever had fresh veg/fruit. & 0 & 0 & 0 & 0 & 0 & 0 \\
\bottomrule
\end{tabular}
}
\caption{Examples from the MFSC dataset with binary labels for  moral sentiment. The examples show the labels provided by 6 out of 24 annotators.}
\label{tab:example_annotations}
\end{table}



\noindent \textbf{Brexit Hate dataset:}
\noindent Hate speech detection has become one of the primary subjective tasks studied in the NLP community \citep{akhtar2019new, sang2022origin, sap2021annotators}. The Brexit Hate dataset (Brexit) introduced by \citet{akhtar2021whose}, consists of 1,120 English tweets collected with keywords related to immigration and Brexit. The dataset was annotated with hate speech (in particular xenophobia and islamophobia), aggressiveness, offensiveness, and stereotype, by six annotators belonging to two distinct groups: a target group of three Muslim immigrants in the UK, and a control group who were researchers with Western background. For our experiments, we use the overall hate label. 

\autoref{tab-app:data_stats} provides the datasets' statistics, including Fleiss's kappa \citep{fleiss1971measuring}, which measures the inter-annotator agreement. The low agreement values highlight the subjective nature of these tasks.
Furthermore, the '\%Pos.' column in \autoref{tab-app:data_stats} shows the class imbalance in \emnlp{the Brexit dataset} and the scarcity of positive class annotations. For example, in the Brexit dataset, only 12\% of samples, on average, were labeled as "Hate".

\renewcommand{\arraystretch}{1.3}
\setlength{\tabcolsep}{4pt}
\begin{table}[h!]
\centering
\small
\begin{tabular}{@{}lcccc@{}}
\toprule
Dataset & Size & $|\mathcal{A}|$ & Kappa & $\%$Pos. \\
\midrule
Brexit & $1120$ & $6$ & $0.34$ & $12.86$ \\
MFSC (Moral) & $2000$ & $24$ & $0.26$ & $63.69$ \\
\bottomrule
\end{tabular}
\caption{Statistics of the datasets used in our experiments. $|\mathcal{A}|$ denotes the number of annotators, Kappa represents Fleiss's kappa inter-annotator agreement, and $\%$Pos. indicates the average percentage of positive class annotations across annotators.}
\label{tab-app:data_stats}
\end{table}



\subsection{Experiment Setup} \label{sec:exp-setup}
We designed our experiments to study the impact of each component of the framework towards our two objectives: maximizing average performance and minimizing annotation budget. 

We use multitask learning (MTL) on all the annotators as our baseline and assess the efficacy of our framework compared to this baseline in capturing individual annotators' perspectives under a range of budget constraints. Specifically, for our approach, we vary the budget $B$ by changing the size of $| \mathcal{A}_{mtl}|$. Recall that $B = |D| = \sum |D_{a_i}|$ and $|D^{fs}_{a_i}| = k << |D_{a_i}|$. Also, recall that under our proposed framework the annotators $\mathcal{A}$ are divided into two sets $ \mathcal{A}_{mtl}$ and $ \mathcal{A}_{fs}$. Since the cost of annotating a few samples per new annotator is negligible ($\frac{|D^{fs}_{a_i}|}{|D_{a_i}|}$ is close to 0) the budget under our proposed framework can be reduced to 
 \begin{align*}
 & B_{ours}   \approx \sum_{a_i \in \mathcal{A}_{\text{mtl}}} |D_{a_i}| \\
 & =\frac{\sum_{a_i \in \mathcal{A}_{\text{mtl}}} |D_{a_i}|}{\sum_{a_i \in \mathcal{A}} |D_{a_i}|} \times B 
  = \frac{|\mathcal{A}_{mtl}| }{ |\mathcal{A}| } \times B
 \end{align*}


For example, the \mf dataset has $|\mathcal{A}|=24$ annotators. Hence, $25\% B$ shows the scenarios where $|\mathcal{A}_{mtl}| = 6$. 
Whereas, for the baseline, we vary the budget $B$ by changing the size of $D_{a_i}$ for all annotators.
In the given example, a $25\% B$ for the baseline means using only $25\%$ of $D_{a_i}$ for each $a_i$.

To ensure that our results are not driven by the specific choices of $\mathcal{A}_{mtl}$, we run our experiments for each budget on multiple samples of $\mathcal{A}_{mtl} \subset \mathcal{A}$. Specifically, we run our models with all possible choices of $\mathcal{A}_{mtl}$ for Brexit dataset and 20 different samples of $\mathcal{A}_{mtl}$ for the \mf dataset.

For each annotator $a_i$, $F^{a_i}_1$ denotes the performance on predicting $a_i$'s labels. We use $F^{fs}_1$ and  $F^{mtl}_1$ to denote the average of $F^{a_i}_1$ scores when $a_i \in \mathcal{A}_{fs}$ and $a_i \in \mathcal{A}_{mtl}$ respectively.
For our framework, we also calculate the overall performance for all annotators $F^{\text{overall}}_1$ as the weighted average of $F^{fs}_1$ and $F^{mtl}_1$. 
\subsection{Implementation Details} \label{implementation}
We evaluate our framework using three base models: RoBERTa (both base and large versions) \citep{liu2019roberta}, and Llama-3 \citep{touvron2023llama}.

\noindent\textbf{Roberta Models}: All multitask models undergo hyperparameter tuning for learning rate and weight decay (see Appendix \ref{sec:app-imp_details}) and are trained for 5 epochs. The best model is selected based on the validation $F_1$ score, and its optimal hyperparameters are also applied in the few-shot stage. All models converge within 5 epochs for MTL and 50 epochs for few-shot learning. To ensure robustness, experiments are repeated with three random seeds.

\noindent\textbf{Llama-3}: We use Llama-3-8B and employ LoRA \citep[Low-Rank Adaptation;][]{hu2021lora} for fine-tuning. We conduct hyperparameter tuning for LoRA parameters, in addition to learning rate and weight decay. In the second stage of our framework, we use the same set of hyperparameters determined in the first stage for few-shot adaptation. The hyperparameters used in our MTL training, along with other variables, are shown in \autoref{tab:hyperparameters}.

For all models, we use the \textit{AdamW} optimizer. For the Brexit dataset, we utilize predefined train, validation, and test splits provided within the dataset\footnote{\url{https://le-wi-di.github.io/}}, and we employ a weighted random sampler to account for the imbalance in the labels of each annotator. For the MFSC dataset, we allocate 80\% for training, 10\% for validation, and the remaining 10\% for testing. 

In the few-shot stage, we experiment with four different values of $k$ (16, 32, 64, and 128). We report the results for $k = 128$ in the next section, while the experiments for other values of $k$ are provided in Appendix \ref{app-result_details}.


\section{Results and Analysis} 


\subsection{Towards Better Performance with Less Annotation Budget} \label{sec:res_main}
\begin{figure*}
\centering
\begin{subfigure}[b]{\textwidth}
   \centering
   \includegraphics[width=0.92\linewidth]{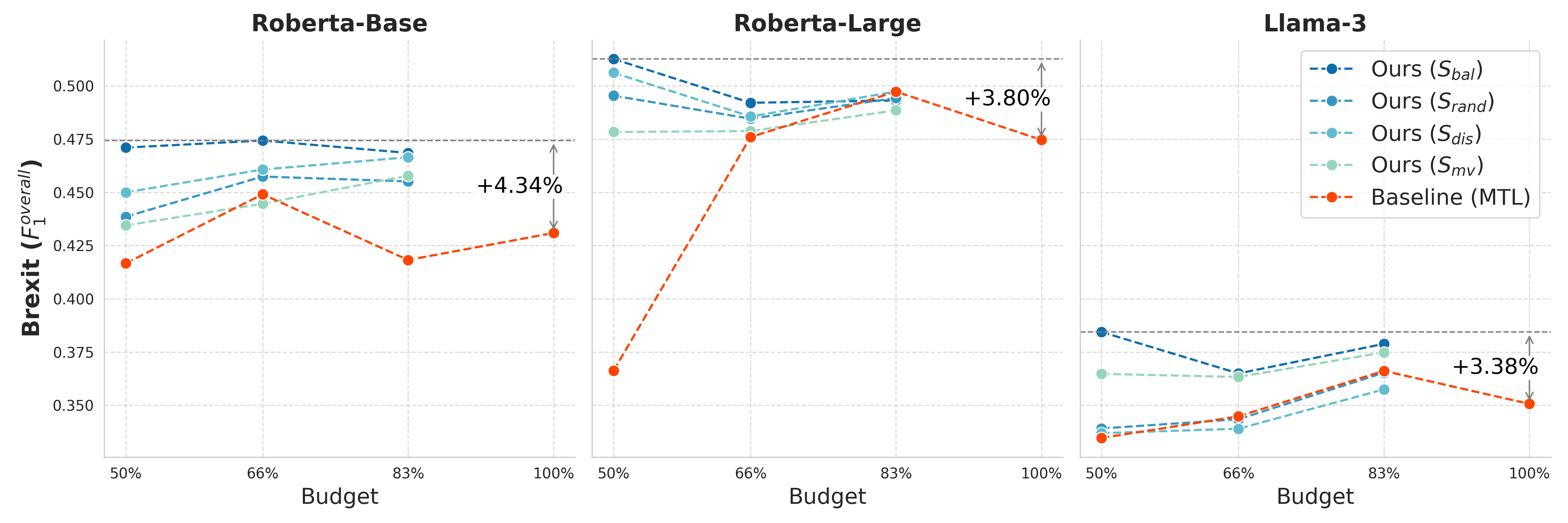}
\end{subfigure}
\begin{subfigure}[b]{\textwidth}
    \centering
   \includegraphics[width=0.92\linewidth]{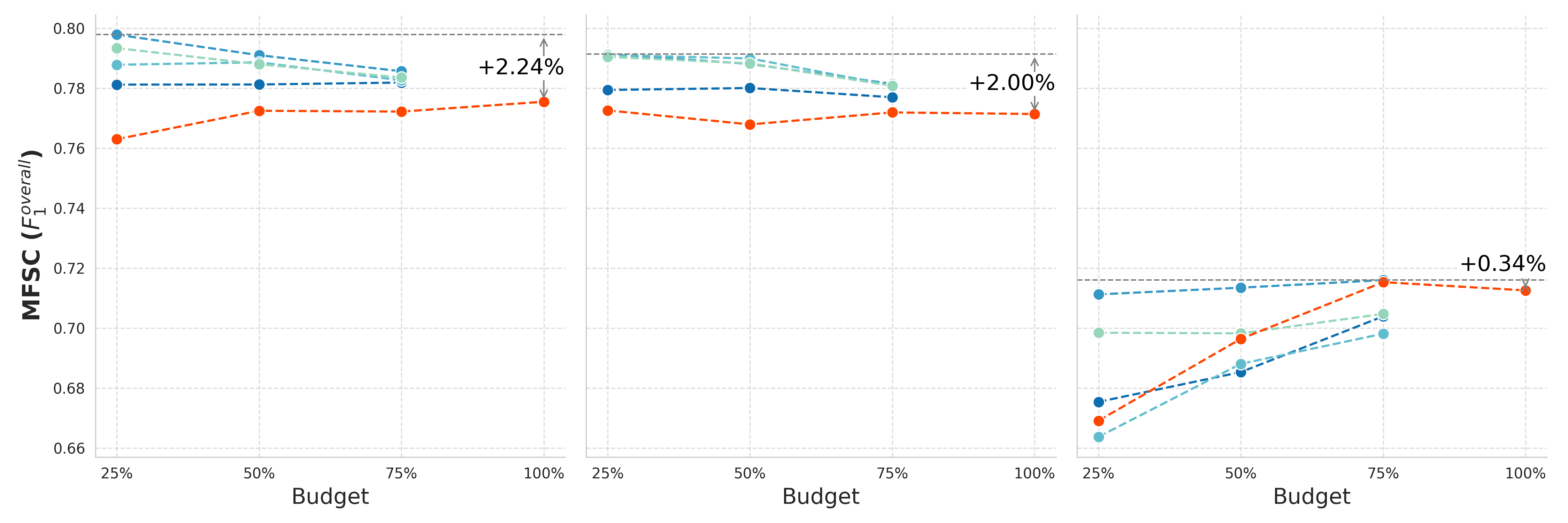}
\end{subfigure}
\caption{Overall $F_1$ score ($F_1^{overall}$) of our framework compared to the baseline across all three base models on both datasets. We observe a \textbf{3.8\%} performance gain with only 50\% of the annotation budget on Brexit dataset, and \textbf{ 2.24\%} gain with 25\% of the annotation budget on MFSC dataset, on the best performing base models.}
\label{fig:all_overall_results}
\end{figure*}

\autoref{fig:all_overall_results} shows the overall $F_1$ scores of our framework for two datasets across varying budgets, evaluated using three different base models. We observe that our framework consistently outperforms the baseline, particularly at lower budget levels. More importantly, our method surpasses the baseline trained with 100\% of the budget using as little as 25\% of the original budget across all three base models, demonstrating its model-agnostic efficacy.

Specifically, at the lowest budget level in the Brexit dataset, our framework with balanced sampling ($\mathcal{S}_{bal}$) achieves performance gains of 5\%, 14\%, and 5\% compared to the baseline when trained with RoBERTa-Base, RoBERTa-Large, and Llama-3, respectively. 
Compared to the full budget, our method shows a gain of 3.8\% with RoBERTa-Large and 3.38\% with Llama-3 using only 50\% of the original budget, and a gain of 4.34\% with RoBERTa-Base using 66\% of the budget.

For the MFSC dataset, our framework, regardless of the sampling method, outperforms the baseline across all budget levels with the RoBERTa-Base and RoBERTa-Large models. Additionally, with the Llama-3 model at 25\% of the budget, our method has 4\% gain compared to baseline. 

These findings demonstrate the success of our framework in achieving its dual objectives: enhancing performance across all annotators while reducing annotation budget requirements. We also conduct an ablation study by omitting the first MTL stage and employing random few-shot sampling for each annotator. Additionally, we compare our framework with more baselines (see Appendix \ref{sec:app-ablation}). 

\begin{figure*}
\begin{subfigure}[b]{\textwidth}
   \centering
   \includegraphics[width=0.92\linewidth]{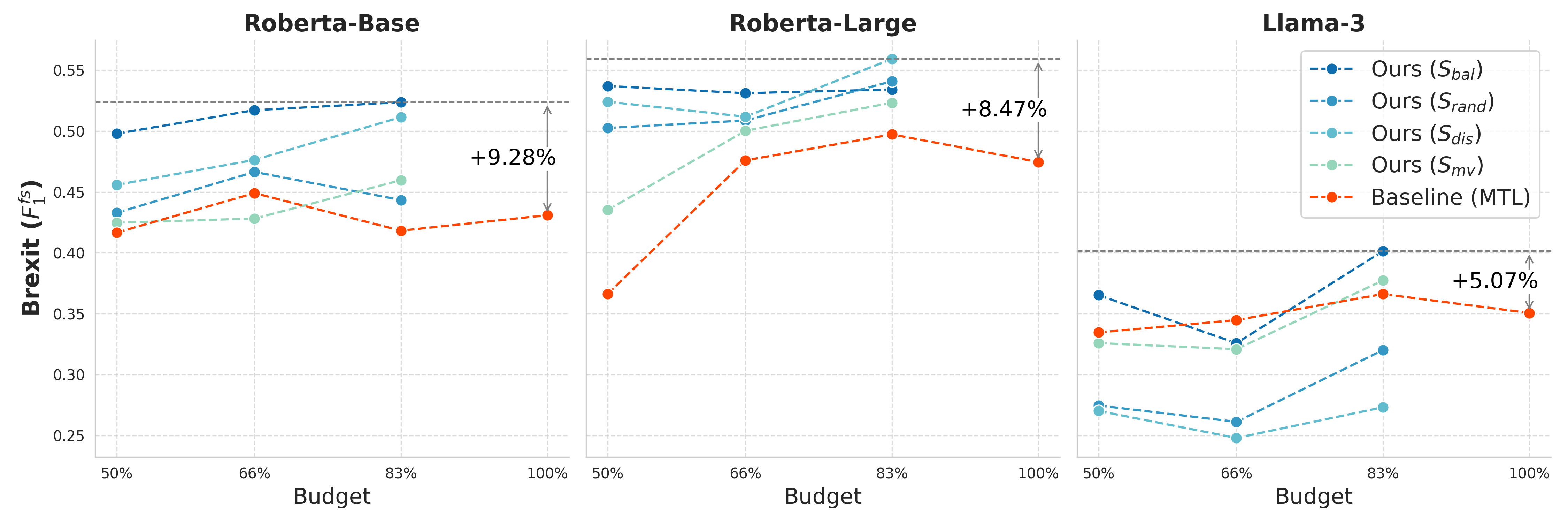}
\end{subfigure}
\begin{subfigure}[b]{\textwidth}
    \centering
   \includegraphics[width=0.92\linewidth]{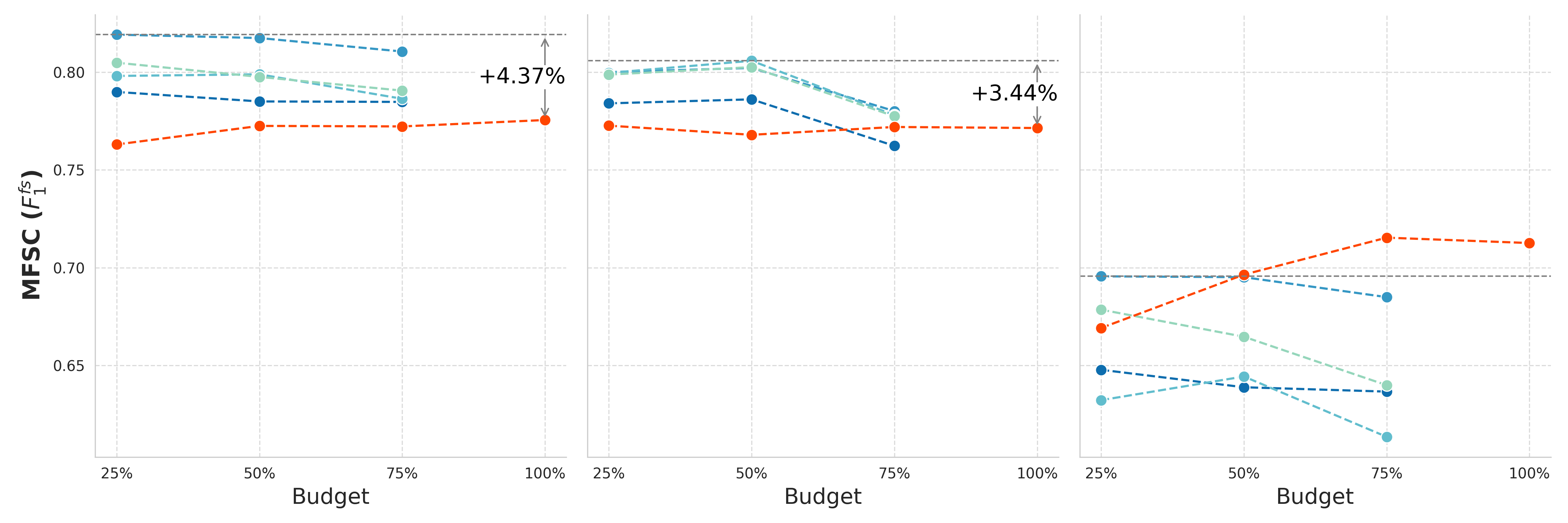}
\end{subfigure}
\caption{Few-shot $F_1$ score ($F_1^{fs}$) of our framework compared to the baseline across all three base models on both datasets. We observe a \textbf{8.47\%} performance gain with  83\% of the annotation budget on Brexit dataset, and \textbf{4.37\% }gain with 25\% of the annotation budget on MFSC dataset, on the best performing base models.}
\label{fig:all_fs_results}
\label{fig:all_results}
\end{figure*}

\noindent\textbf{Incorporating a New Annotator}:
The second stage of our framework suggests that few-shot adaptation not only allows us to integrate a new annotator into an already existing model with minimal budget, but also maintains the annotator's performance. To validate this ability, in \autoref{fig:all_fs_results} we compare the performance of the second stage of our framework ($F_1^{fs}$) with the baseline.

For the Brexit dataset, $F_1^{fs}$ scores exceed the baseline across all base models by 9.28\%, 8.47\%, and 5.07\%, respectively. The balanced sampling ($\mathcal{S}_{bal}$) method consistently performs well across all models. Similarly, for the MFSC dataset, our framework achieves higher $F_1^{fs}$ scores regardless of sampling method, except with Llama-3 model.

Overall, our results on both datasets show that the few-shot stage of our framework results in models that outperform the multitask learning baseline.

\noindent\textbf{Base Model Comparison}: Generally, the RoBERTa models perform better than the Llama-3 model in MTL setting. Llama-3 model, despite undergoing the most hyperparameter search and utilizing the most GPU hours to find optimal parameters, still performs significantly poorer than the other two models, especially when fine-tuned in a few-shot setting. A potential reason for this disparity is that larger models like Llama-3, while generally more capable, require extensive hyperparameter tuning to optimize their performance.  Additionally, they have stronger biases, making it more challenging to adapt them to different perspectives \citep{naveed2023comprehensive,liu2023survey}.

\subsection{Reduced Performance Disparities across Annotators} \label{sec:dispar}

Ensuring a comprehensive representation of annotators' viewpoints is crucial in modeling subjective tasks. To achieve this goal, a critical criterion is to create models that not only improve the aggregated performance but also demonstrate fair and equitable performance across all annotators. For example, if the $F_1$ scores of one model for two annotators are 0.6 and 0.8, respectively, while the second model scores 0.7 for both annotators, the latter is considered a better model. Although the average performance is the same for both models, the first model has a disparate negative impact on the first annotator.
This is important because performance disparities among social groups (in our case annotators) can lead to biased models, limiting the system's ability to accurately reflect diverse perspectives and potentially perpetuating inequalities in the outputs of subjective tasks \citep{buolamwini2018gender}.
Merely relying on aggregated performance measures, such as the average across all annotators, fails to provide a comprehensive understanding of how well the model captures the varying perspectives of different annotators. For instance, it remains unclear whether the average performance improves because the approach better captures the perspectives of all the annotators or only a subset of them.
Hence, we look into the standard deviation of performance across all annotators as a measure of performance disparity: $d = \sqrt{\frac{1}{N-1} \sum_{i=1}^N (F^{a_i}_1 - \overline{F^{overall}_1})^2}$. Lower standard deviations indicate more equitable models.

\begin{figure}[ht]
    \centering
    \includegraphics[width=\columnwidth]{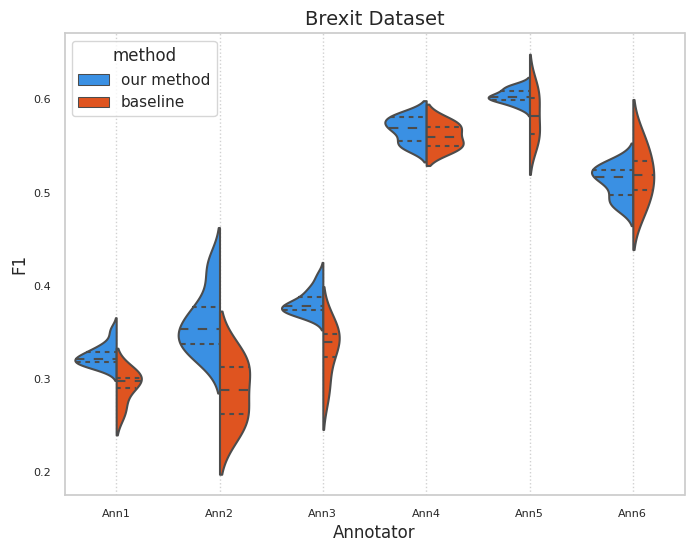}
    \caption{Comparison of Annotator level $F_1$ scores ($F_1^{a_i}$) on the Brexit dataset between MTL model and our framework, leveraging the $\mathcal{S}_{bal}$ sampling method for all budgets and shots on RoBERTa-base model}
    \label{fig:brexit-fairness} 
\end{figure}

As shown in \autoref{tab:variance}, our approach results in lower performance disparities ($d$) compared to the MTL baseline regardless of the base model, across all budgets for the MFSC dataset. For the Brexit dataset, this improvement is observed at lower budgets (50\% and 66\%).
Among the various sampling strategies, the balanced sampling strategy ($S_{bal}$) consistently results in lower $d$ for MFSC dataset. 
When comparing different base models, the lowest $d$ is achieved using RoBERTa-Base model. Specifically, for the MFSC dataset, there is a 1.1\% reduction in $d$ at 25\% of the budget compared to the baseline, and for the Brexit dataset, there is a 7.5\% reduction in $d$ at 50\% of the budget.
\autoref{fig:brexit-fairness} visualizes this model's performance in comparison to the MTL baseline for each annotator. Notably, our framework improves performance for annotators in the non-Western control group (i.e., the first three annotators) while maintaining the performance of the remaining annotators.

Overall, these findings suggest that our proposed framework not only improves the overall performance of all annotators but also yields models that are more fair and equitable. 
\renewcommand{\arraystretch}{1.5}
\setlength{\tabcolsep}{2pt}
\begin{table}
\small
\centering
\begin{tabular}{ll|llll|llll}
\Xhline{3\arrayrulewidth}
 \multicolumn{2}{c|}{\multirow{2}{*}{$d \downarrow$}} & \multicolumn{4}{c|}{Brexit} & \multicolumn{4}{c}{\mf} \\  
\multicolumn{2}{c|}{} & $50\%$ &    $66\%$ & $83\%$ & $100\%$ & $25\%$ & $50\%$ & $75\%$ & $100\%$ \\ \Xhline{2\arrayrulewidth}
\multirow{5}{*}{\rotatebox[origin=c]{90}{Roberta-Base}} & MTL & .168 & .139 & .131 & .130 & .128 & .136 & .127 & .130 \\
 & $S_{bal}$ & \textbf{.093} & \textbf{.108} & \textbf{.117} & \cellcolor{lightgray} & \textbf{.117} & \textbf{.122} & \textbf{.121} & \cellcolor{lightgray} \\
 & $S_{dis}$ & .111 & .120 & .124 & \cellcolor{lightgray} & .130 & .129 & .127 & \cellcolor{lightgray} \\
 & $S_{mv}$ & .137 & .142 & .132 & \cellcolor{lightgray} & .126 & .128 & .127 & \cellcolor{lightgray} \\
 & $S_{rand}$ & .131 & .127 & .136 & \cellcolor{lightgray} & .134 & .133 & .128 & \cellcolor{lightgray} \\
\cline{1-10}
\multirow{5}{*}{\rotatebox[origin=c]{90}{Roberta-Large}} 
& MTL & .170 & .136 & \textbf{.117} & .155 & .152 & .143 & .146 & .149 \\
 & $S_{bal}$ & \textbf{.102} & \textbf{.127} & .148 & \cellcolor{lightgray} & \textbf{.117} & \textbf{.121} & \textbf{.127} & \cellcolor{lightgray} \\
 & $S_{dis}$ & .112 & .134 & .140 & \cellcolor{lightgray} & .132 & .128 & .133 & \cellcolor{lightgray} \\
 & $S_{mv}$ & .134 & .141 & .149 & \cellcolor{lightgray} & .130 & .128 & .131 & \cellcolor{lightgray} \\
 & $S_{rand}$ & .120 & .131 & .146 & \cellcolor{lightgray} & .127 & .129 & .130 & \cellcolor{lightgray} \\
\cline{1-10}
\multirow{5}{*}{\rotatebox[origin=c]{90}{Llama-3}} & MTL & .172 & .122 & \textbf{.112} & .130 & .189 & .171 & .167 & .158 \\
 & $S_{bal}$ & .117 & .113 & .120 & \cellcolor{lightgray} & \textbf{.137} & \textbf{.147} & \textbf{.158} & \cellcolor{lightgray} \\
 & $S_{dis}$ & \textbf{.114} & .113 & .120 & \cellcolor{lightgray} & .198 & .187 & .176 & \cellcolor{lightgray} \\
 & $S_{mv}$ & .127 & \textbf{.103} & .124 & \cellcolor{lightgray} & .170 & .174 & .170 & \cellcolor{lightgray} \\
 & $S_{rand}$ & .150 & .128 & .119 & \cellcolor{lightgray} & .176 & .173 & .166 & \cellcolor{lightgray} \\
\cline{1-10}
\bottomrule
\end{tabular}
\caption{Performance disparities across annotators ($d \downarrow$). The best values are shown in bold. $\mathcal{S}_{bal}$, $\mathcal{S}_{dis}$, $\mathcal{S}_{mv}$, and $\mathcal{S}_{rand}$ refer to the sampling functions used in the second stage of our framework (\S \ref{sec:method}). We generally observe lower performance disparities with our framework compared to the MTL baseline.
}
\label{tab:variance}
\end{table}

\subsection{Annotator-level Analysis} \label{sec:ann_level}

Here, we delve into the relationship between annotator-level variables. Recall that our framework is trained on $\mathcal{A}_{mtl}$ in the initial stage, followed by fine-tuning for each $a \in \mathcal{A}_{fs}$. Hence, a practical question arises: does the choice of the set $\mathcal{A}_{mtl}$ matter? In other words, would the similarity or divergence in perspectives among annotators in this set impact the performance on $\mathcal{A}_{fs}$? Investigating this is crucial, as identifying such an effect would necessitate a thoughtful selection of $\mathcal{A}_{mtl}$. To examine this, we conduct the following analysis:

\noindent Disagreement within $\mathcal{A}_{mtl}$ and performance on $\mathcal{A}_{fs}$:
The aim of this analysis is to investigate whether there is a relationship between the disagreement within annotators in  $\mathcal{A}_{mtl}$ and the performance of the newly adopted annotators in $A_{fs}$.

To test this relationship, we employ a mixed-effects model to predict the performance of $a \in \mathcal{A}_{fs}$ by the  \textit{agreement within $\mathcal{A}_{mtl}$} denoted as $d^1$ \citep{fleiss1971measuring}. The model controls for $k$, budget $B$, \textit{and agreement between $\mathcal{A}_{fs}$ and $\mathcal{A}_{mtl}$}, denoted using $d^2$, incorporating random effects for $\mathcal{A}_{mtl}$ and $\mathcal{A}_{fs}$. The formula for this model is as follows:
\begin{equation}
\begin{split}
f_{ij} = & \beta_0 + \beta_1 d^1_{j} + \beta_2 k_{ij} + \beta_3 B_{j} \\
& + \beta_4 d^2_{ij} + u_{0i} + v_{1j} + e_{ij}
\end{split}
 \label{eq:5}
\end{equation}
where $f_{ij}$ denotes the performance of $i^{\text{th}}$ annotator in $A_{fs}$ on the model trained on a $j^{\text{th}}$ sample of $A_{mtl}$. The fixed effects coefficients are represented by  $\beta_0$ to $\beta_4$, and the random effects for $i$ and $j$ are represented by $u_{0i}$, $v_{1j}$ respectively. $e_{ij}$  denotes the residual error term.
To see the impact of sampling strategies, we run a total of four models, each corresponding to the performance results obtained from one of the strategies  ($\mathcal{S}_{bal}, \mathcal{S}_{dis}, \mathcal{S}_{mv}, \mathcal{S}_{rand}$).

The findings regarding Brexit indicate no statistically significant effect of agreement within $\mathcal{A}_{mtl}$ ($d^1$) on the performance.
For the \mf dataset, a significant effect was observed only for results obtained from $\mathcal{S}_{bal}$ 
($\beta_1 = -0.052, SE =  0.012,  p < 0.001$).
This implies that a unit decrease in $d^1$, corresponding to moving from full agreement to full disagreement, is associated with a 0.052 increase in the $F_1$ score. This finding suggests that selecting a diverse $\mathcal{A}_{mtl}$ with high disagreement can potentially be advantageous.

\section{Conclusion}
We introduced a framework for annotation collection and annotator modeling in subjective tasks. Our framework aims to minimize the annotation budget required to model a fixed number of annotators while maximizing the predictive performance for each annotator. Our approach involves collecting annotations from an initial set of annotators and building a multitask model that captures general task patterns while signaling differences among individual annotators. Subsequently, we utilize the annotations from the first stage to select a small set of samples from new annotators that best highlight their deviations from the general patterns. Finally, we use these samples to augment the initial model in a few-shot setting to learn the new annotator's perspective.
We evaluated our framework using three base models, and explored four distinct methods for few-shot sample selection and found that the most effective approach involves balanced and random sample selections.
We introduced a new subjective task dataset Moral Foundations Subjective Corpus (MFSC), of 2000 Reddit posts annotated by 24 annotators for moral sentiment which enabled us to test our framework in scale.
Our experiments on \mf and a hate speech dataset revealed that our framework consistently surpasses previous SOTA with access to as little as 25\% of the original annotation budget. In addition, we showed that our framework yields more equitable models that reduce performance disparities among annotators.
Our cost-efficient framework for subjective tasks allows enhancing the diversity of the captured perspectives, and facilitates the integration of new annotators into pre-existing datasets and models.
\section{Limitations and Ethical Statement}

We acknowledge that the datasets employed in our experiments are not representative of all annotator populations.
While in \mf we recruited a substantial number of annotators and efforts were made to diversify this pool, it is important to note that our sample is limited to undergraduate students at a private university in the US. Consequently, we advocate for the replication and extension of our work with non-student, non-US-based samples.
Furthermore, we exclusively operate with English data and focus on datasets related to moral sentiment prediction and hate speech detection tasks. This may restrict the generalizability of our findings to a broader linguistic and thematic landscape. Despite these constraints, our research lays the groundwork for future research to extend and validate our approach across diverse languages and subjective NLP tasks.
In our experiments, we do not consider the cost of collecting few-shot samples, as discussed in Section \ref{sec:exp-setup}.  We recognize that in certain cases, depending on the  budget and the nature of the task, this assumption can be challenged.  Even with the additional expense of annotating a few samples per new annotator, it is crucial to highlight that our proposed framework substantially reduces annotation cost,  especially as the number of included perspectives grows.

In the \mf dataset the annotators  underwent four sessions of training, including guidance on avoiding potential adverse consequences of annotations, and were compensated at a rate of \$17 per hour. The study protocol received approval from the Institutional Review Board (IRB), and all annotators consented to both the terms outlined in an information sheet provided by the IRB about the study and the sharing of their responses to the psychological questionnaires along with their annotations. We emphasize that \mf is created with the intention of exploring subjectivity and different perspectives in this context and it should not be used for any other purposes. 

\bibliography{anthology,main}

\begin{thebibliography}{38}
\expandafter\ifx\csname natexlab\endcsname\relax\def\natexlab#1{#1}\fi

\bibitem[{Akhtar et~al.(2019)Akhtar, Basile, and Patti}]{akhtar2019new}
Sohail Akhtar, Valerio Basile, and Viviana Patti. 2019.
\newblock A new measure of polarization in the annotation of hate speech.
\newblock In \emph{International Conference of the Italian Association for Artificial Intelligence}, pages 588--603. Springer.

\bibitem[{Akhtar et~al.(2020)Akhtar, Basile, and Patti}]{akhtar2020modeling}
Sohail Akhtar, Valerio Basile, and Viviana Patti. 2020.
\newblock Modeling annotator perspective and polarized opinions to improve hate speech detection.
\newblock In \emph{Proceedings of the AAAI Conference on Human Computation and Crowdsourcing}, volume~8, pages 151--154.

\bibitem[{Akhtar et~al.(2021)Akhtar, Basile, and Patti}]{akhtar2021whose}
Sohail Akhtar, Valerio Basile, and Viviana Patti. 2021.
\newblock Whose opinions matter? perspective-aware models to identify opinions of hate speech victims in abusive language detection.
\newblock \emph{arXiv preprint arXiv:2106.15896}.

\bibitem[{Atari et~al.(2023)Atari, Haidt, Graham, Koleva, Stevens, and Dehghani}]{atari2023morality}
Mohammad Atari, Jonathan Haidt, Jesse Graham, Sena Koleva, Sean~T Stevens, and Morteza Dehghani. 2023.
\newblock Morality beyond the weird: How the nomological network of morality varies across cultures.
\newblock \emph{Journal of Personality and Social Psychology}.

\bibitem[{Basile(2020)}]{basile2020s}
Valerio Basile. 2020.
\newblock It’s the end of the gold standard as we know it. on the impact of pre-aggregation on the evaluation of highly subjective tasks.
\newblock In \emph{2020 AIxIA Discussion Papers Workshop, AIxIA 2020 DP}, volume 2776, pages 31--40. CEUR-WS.

\bibitem[{Basile et~al.(2021)Basile, Cabitza, Campagner, and Fell}]{basile2021toward}
Valerio Basile, Federico Cabitza, Andrea Campagner, and Michael Fell. 2021.
\newblock Toward a perspectivist turn in ground truthing for predictive computing.
\newblock \emph{arXiv preprint arXiv:2109.04270}.

\bibitem[{Baumler et~al.(2023)Baumler, Sotnikova, and Daum{\'e}~III}]{baumler2023examples}
Connor Baumler, Anna Sotnikova, and Hal Daum{\'e}~III. 2023.
\newblock Which examples should be multiply annotated? active learning when annotators may disagree.
\newblock In \emph{Findings of the Association for Computational Linguistics: ACL 2023}, pages 10352--10371.

\bibitem[{Buolamwini and Gebru(2018)}]{buolamwini2018gender}
Joy Buolamwini and Timnit Gebru. 2018.
\newblock Gender shades: Intersectional accuracy disparities in commercial gender classification.
\newblock In \emph{Conference on fairness, accountability and transparency}, pages 77--91. PMLR.

\bibitem[{Casola et~al.(2023)Casola, Lo, Basile, Frenda, Cignarella, Patti, and Bosco}]{casola2023confidence}
Silvia Casola, Soda Lo, Valerio Basile, Simona Frenda, Alessandra Cignarella, Viviana Patti, and Cristina Bosco. 2023.
\newblock Confidence-based ensembling of perspective-aware models.
\newblock In \emph{Proceedings of the 2023 Conference on Empirical Methods in Natural Language Processing}, pages 3496--3507.

\bibitem[{Davani et~al.(2023)Davani, Atari, Kennedy, and Dehghani}]{davani2023hate}
Aida~Mostafazadeh Davani, Mohammad Atari, Brendan Kennedy, and Morteza Dehghani. 2023.
\newblock Hate speech classifiers learn normative social stereotypes.
\newblock \emph{Transactions of the Association for Computational Linguistics}, 11:300--319.

\bibitem[{Davani et~al.(2022)Davani, D{\'\i}az, and Prabhakaran}]{davani2022dealing}
Aida~Mostafazadeh Davani, Mark D{\'\i}az, and Vinodkumar Prabhakaran. 2022.
\newblock Dealing with disagreements: Looking beyond the majority vote in subjective annotations.
\newblock \emph{Transactions of the Association for Computational Linguistics}, 10:92--110.

\bibitem[{Deng et~al.(2023)Deng, Zhang, Liu, Wu, Wang, and Mihalcea}]{deng-etal-2023-annotate}
Naihao Deng, Xinliang Zhang, Siyang Liu, Winston Wu, Lu~Wang, and Rada Mihalcea. 2023.
\newblock \href {https://doi.org/10.18653/v1/2023.findings-emnlp.832} {You are what you annotate: Towards better models through annotator representations}.
\newblock In \emph{Findings of the Association for Computational Linguistics: EMNLP 2023}, pages 12475--12498, Singapore. Association for Computational Linguistics.

\bibitem[{D{\'\i}az et~al.(2018)D{\'\i}az, Johnson, Lazar, Piper, and Gergle}]{diaz2018addressing}
Mark D{\'\i}az, Isaac Johnson, Amanda Lazar, Anne~Marie Piper, and Darren Gergle. 2018.
\newblock Addressing age-related bias in sentiment analysis.
\newblock In \emph{Proceedings of the 2018 chi conference on human factors in computing systems}, pages 1--14.

\bibitem[{Ferracane et~al.(2021)Ferracane, Durrett, Li, and Erk}]{ferracane-etal-2021-answer}
Elisa Ferracane, Greg Durrett, Junyi~Jessy Li, and Katrin Erk. 2021.
\newblock \href {https://doi.org/10.18653/v1/2021.naacl-main.129} {Did they answer? subjective acts and intents in conversational discourse}.
\newblock In \emph{Proceedings of the 2021 Conference of the North American Chapter of the Association for Computational Linguistics: Human Language Technologies}, pages 1626--1644, Online. Association for Computational Linguistics.

\bibitem[{Fleiss(1971)}]{fleiss1971measuring}
Joseph~L Fleiss. 1971.
\newblock Measuring nominal scale agreement among many raters.
\newblock \emph{Psychological bulletin}, 76(5):378.

\bibitem[{Garten et~al.(2019)Garten, Kennedy, Hoover, Sagae, and Dehghani}]{garten2019incorporating}
Justin Garten, Brendan Kennedy, Joe Hoover, Kenji Sagae, and Morteza Dehghani. 2019.
\newblock Incorporating demographic embeddings into language understanding.
\newblock \emph{Cognitive science}, 43(1):e12701.

\bibitem[{Graham et~al.(2013)Graham, Haidt, Koleva, Motyl, Iyer, Wojcik, and Ditto}]{graham2013moral}
Jesse Graham, Jonathan Haidt, Sena Koleva, Matt Motyl, Ravi Iyer, Sean~P Wojcik, and Peter~H Ditto. 2013.
\newblock Moral foundations theory: The pragmatic validity of moral pluralism.
\newblock In \emph{Advances in experimental social psychology}, volume~47, pages 55--130. Elsevier.

\bibitem[{Graham et~al.(2016)Graham, Meindl, Beall, Johnson, and Zhang}]{graham2016cultural}
Jesse Graham, Peter Meindl, Erica Beall, Kate~M Johnson, and Li~Zhang. 2016.
\newblock Cultural differences in moral judgment and behavior, across and within societies.
\newblock \emph{Current Opinion in Psychology}, 8:125--130.

\bibitem[{Hu et~al.(2021)Hu, Shen, Wallis, Allen-Zhu, Li, Wang, and Chen}]{hu2021lora}
Edward~J. Hu, Yelong Shen, Phillip Wallis, Zeyuan Allen-Zhu, Yuanzhi Li, Shean Wang, and Weizhu Chen. 2021.
\newblock \href {http://arxiv.org/abs/2106.09685} {Lora: Low-rank adaptation of large language models}.

\bibitem[{Kanclerz et~al.(2022)Kanclerz, Gruza, Karanowski, Bielaniewicz, Mi{\l}kowski, Koco{\'n}, and Kazienko}]{kanclerz2022if}
Kamil Kanclerz, Marcin Gruza, Konrad Karanowski, Julita Bielaniewicz, Piotr Mi{\l}kowski, Jan Koco{\'n}, and Przemyslaw Kazienko. 2022.
\newblock What if ground truth is subjective? personalized deep neural hate speech detection.
\newblock In \emph{Proceedings of the 1st Workshop on Perspectivist Approaches to NLP@ LREC2022}, pages 37--45.

\bibitem[{Kennedy et~al.(2018)Kennedy, Atari, Davani, Yeh, Omrani, Kim, Coombs, Havaldar, Portillo-Wightman, Gonzalez et~al.}]{kennedy2018gab}
Brendan Kennedy, Mohammad Atari, Aida~Mostafazadeh Davani, Leigh Yeh, Ali Omrani, Yehsong Kim, Kris Coombs, Shreya Havaldar, Gwenyth Portillo-Wightman, Elaine Gonzalez, et~al. 2018.
\newblock The gab hate corpus: A collection of 27k posts annotated for hate speech.
\newblock \emph{PsyArXiv. July}, 18.

\bibitem[{Larimore et~al.(2021)Larimore, Kennedy, Haskett, and Arseniev-Koehler}]{larimore2021reconsidering}
Savannah Larimore, Ian Kennedy, Breon Haskett, and Alina Arseniev-Koehler. 2021.
\newblock Reconsidering annotator disagreement about racist language: Noise or signal?
\newblock In \emph{Proceedings of the Ninth International Workshop on Natural Language Processing for Social Media}, pages 81--90.

\bibitem[{Liu et~al.(2019)Liu, Ott, Goyal, Du, Joshi, Chen, Levy, Lewis, Zettlemoyer, and Stoyanov}]{liu2019roberta}
Yinhan Liu, Myle Ott, Naman Goyal, Jingfei Du, Mandar Joshi, Danqi Chen, Omer Levy, Mike Lewis, Luke Zettlemoyer, and Veselin Stoyanov. 2019.
\newblock Roberta: A robustly optimized bert pretraining approach.
\newblock \emph{arXiv preprint arXiv:1907.11692}.

\bibitem[{Liu et~al.(2023)}]{liu2023survey}
Zheng Liu et~al. 2023.
\newblock A survey of large language models.
\newblock \emph{arXiv preprint arXiv:2303.18223}.

\bibitem[{Naveed et~al.(2023)Naveed, Khan, Qiu, Saqib, Anwar, Usman, Barnes, and Mian}]{naveed2023comprehensive}
Humza Naveed, Asad~Ullah Khan, Shi Qiu, Muhammad Saqib, Saeed Anwar, Muhammad Usman, Nick Barnes, and Ajmal Mian. 2023.
\newblock A comprehensive overview of large language models.
\newblock \emph{arXiv preprint arXiv:2307.06435}.

\bibitem[{Pavlick and Kwiatkowski(2019)}]{pavlick2019inherent}
Ellie Pavlick and Tom Kwiatkowski. 2019.
\newblock Inherent disagreements in human textual inferences.
\newblock \emph{Transactions of the Association for Computational Linguistics}, 7:677--694.

\bibitem[{Peterson et~al.(2019)Peterson, Battleday, Griffiths, and Russakovsky}]{peterson2019human}
Joshua~C Peterson, Ruairidh~M Battleday, Thomas~L Griffiths, and Olga Russakovsky. 2019.
\newblock Human uncertainty makes classification more robust.
\newblock In \emph{Proceedings of the IEEE/CVF international conference on computer vision}, pages 9617--9626.

\bibitem[{Plank(2022)}]{plank-2022-problem}
Barbara Plank. 2022.
\newblock \href {https://aclanthology.org/2022.emnlp-main.731} {The {``}problem{''} of human label variation: On ground truth in data, modeling and evaluation}.
\newblock In \emph{Proceedings of the 2022 Conference on Empirical Methods in Natural Language Processing}, pages 10671--10682, Abu Dhabi, United Arab Emirates. Association for Computational Linguistics.

\bibitem[{Plank et~al.(2014)Plank, Hovy, and S{\o}gaard}]{plank2014linguistically}
Barbara Plank, Dirk Hovy, and Anders S{\o}gaard. 2014.
\newblock Linguistically debatable or just plain wrong?
\newblock In \emph{Proceedings of the 52nd Annual Meeting of the Association for Computational Linguistics (Volume 2: Short Papers)}, pages 507--511.

\bibitem[{Prabhakaran et~al.(2021)Prabhakaran, Davani, and Diaz}]{prabhakaran2021releasing}
Vinodkumar Prabhakaran, Aida~Mostafazadeh Davani, and Mark Diaz. 2021.
\newblock On releasing annotator-level labels and information in datasets.
\newblock \emph{arXiv preprint arXiv:2110.05699}.

\bibitem[{Sang and Stanton(2022)}]{sang2022origin}
Yisi Sang and Jeffrey Stanton. 2022.
\newblock The origin and value of disagreement among data labelers: A case study of individual differences in hate speech annotation.
\newblock In \emph{International Conference on Information}, pages 425--444. Springer.

\bibitem[{Sap et~al.(2021)Sap, Swayamdipta, Vianna, Zhou, Choi, and Smith}]{sap2021annotators}
Maarten Sap, Swabha Swayamdipta, Laura Vianna, Xuhui Zhou, Yejin Choi, and Noah~A Smith. 2021.
\newblock Annotators with attitudes: How annotator beliefs and identities bias toxic language detection.
\newblock \emph{arXiv preprint arXiv:2111.07997}.

\bibitem[{Soto and John(2017)}]{soto2017short}
Christopher~J Soto and Oliver~P John. 2017.
\newblock Short and extra-short forms of the big five inventory--2: The bfi-2-s and bfi-2-xs.
\newblock \emph{Journal of Research in Personality}, 68:69--81.

\bibitem[{Talat and Hovy(2016)}]{waseem2016hateful}
Zeerak Talat and Dirk Hovy. 2016.
\newblock \href {https://www.aclweb.org/anthology/N16-2013.pdf} {Hateful symbols or hateful people? predictive features for hate speech detection on twitter}.
\newblock In \emph{Proceedings of the NAACL student research workshop}, pages 88--93.

\bibitem[{Touvron et~al.(2023)Touvron, Lavril, Izacard, Martinet, Lachaux, Lacroix, Rozière, Goyal, Hambro, Azhar et~al.}]{touvron2023llama}
Hugo Touvron, Thibaut Lavril, Gautier Izacard, Xavier Martinet, Marie-Anne Lachaux, Timothée Lacroix, Baptiste Rozière, Naman Goyal, Eric Hambro, Faisal Azhar, et~al. 2023.
\newblock Llama: Open and efficient foundation language models.
\newblock \emph{arXiv preprint arXiv:2302.13971}.

\bibitem[{Trager et~al.(2022)Trager, Ziabari, Davani, Golazazian, Karimi-Malekabadi, Omrani, Li, Kennedy, Reimer, Reyes et~al.}]{trager2022moral}
Jackson Trager, Alireza~S Ziabari, Aida~Mostafazadeh Davani, Preni Golazazian, Farzan Karimi-Malekabadi, Ali Omrani, Zhihe Li, Brendan Kennedy, Nils~Karl Reimer, Melissa Reyes, et~al. 2022.
\newblock The moral foundations reddit corpus.
\newblock \emph{arXiv preprint arXiv:2208.05545}.

\bibitem[{Uma et~al.(2021)Uma, Fornaciari, Hovy, Paun, Plank, and Poesio}]{uma2021learning}
Alexandra~N Uma, Tommaso Fornaciari, Dirk Hovy, Silviu Paun, Barbara Plank, and Massimo Poesio. 2021.
\newblock Learning from disagreement: A survey.
\newblock \emph{Journal of Artificial Intelligence Research}, 72:1385--1470.

\bibitem[{Wang and Plank(2023)}]{wang2023actor}
Xinpeng Wang and Barbara Plank. 2023.
\newblock Actor: Active learning with annotator-specific classification heads to embrace human label variation.
\newblock \emph{arXiv preprint arXiv:2310.14979}.

\end{thebibliography}
\bibliographystyle{acl_natbib}

\appendix

\section{Dataset Details} \label{sec:app-data-stats}
\begin{figure*}[ht]
    \centering
    \includegraphics[width=\textwidth]{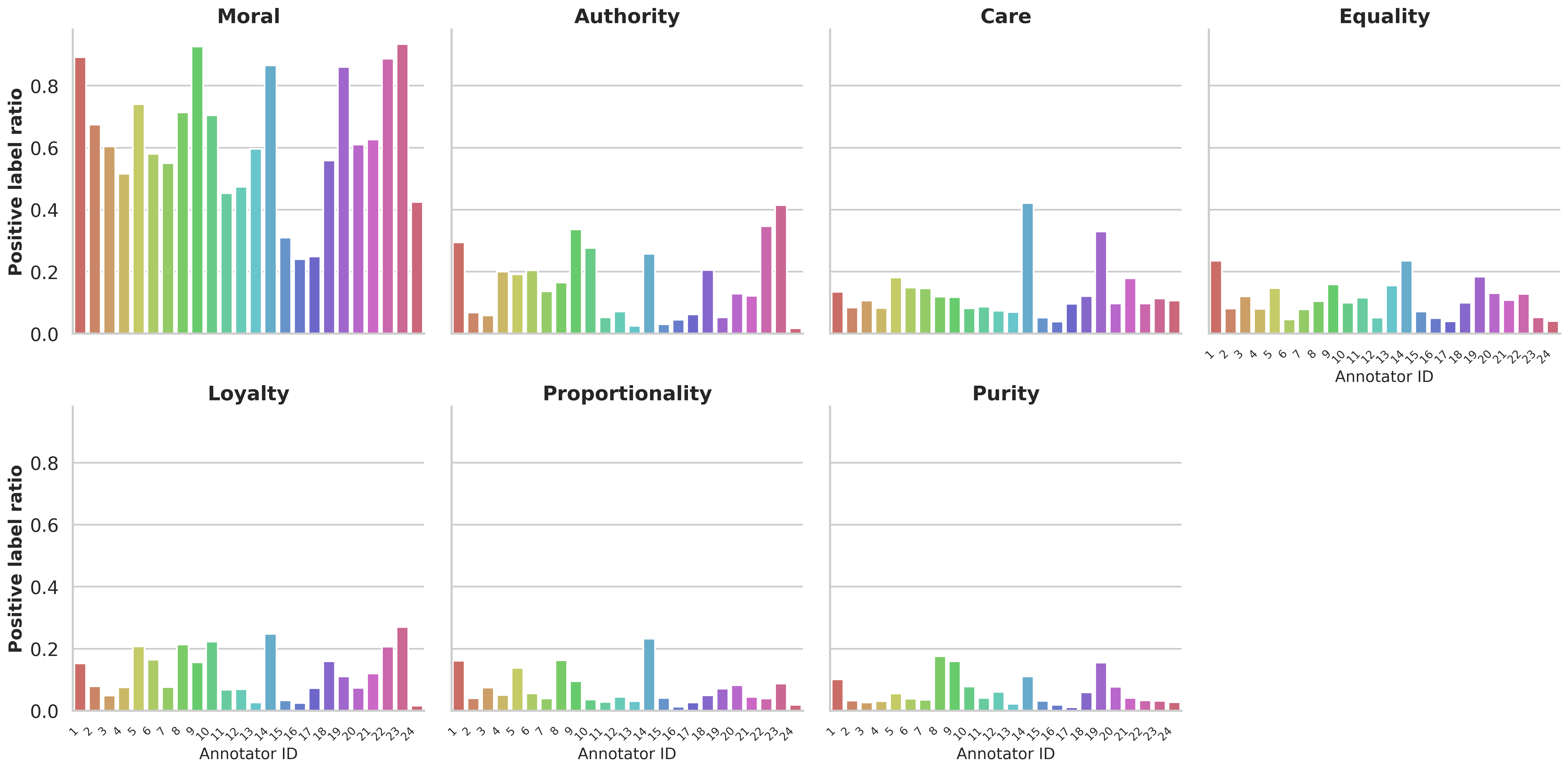}
    \caption{Distribution of the labels across annotators in MFSC dataset}
    \label{fig:mfrc_label_dist}
\end{figure*}

Each sample in the MFSC dataset is annotated with a binary label indicating moral sentiment: 1 if the sentence pertains to morality and 0 if it does not. Additionally, we have collected more fine-grained labels of morality (i.e., Purity, Harm, Loyalty, Authority, Proportionality, or Equality) following the procedure outlined by \citet{trager2022moral}. Specifically, if a sample is labeled as moral, the annotator can select any of the applicable moral categories for that text. The distribution of these labels across annotators is demonstrated in \autoref{fig:mfrc_label_dist}. The dataset also includes additional metadata information, such as confidence levels for each instance using a 3-level measure (\textit{confident}, \textit{somewhat confident}, and \textit{not confident}). Furthermore, we collected annotator responses for the ``Big Five Inventory-2-Short'' questionnaire \citep{soto2017short}. The MFSC dataset provides an opportunity to explore the subjective nature of morality. The substantial number of annotators, along with their questionnaire responses, enables future researchers to investigate the modeling of subjective tasks on a larger scale.
See \autoref{tab:example_annotations} and \autoref{tab:brexit_examples} for sample annotations for MFSC and Brexit datasets.


\setlength{\tabcolsep}{4pt}

\begin{table}[ht]
\centering
\resizebox{\columnwidth}{!}{
\begin{tabular}{@{}p{8.5cm} c c c c c c@{}}
\toprule
\textbf{Brexit examples} & \textbf{a$_1$} & \textbf{a$_2$} & \textbf{a$_3$} & \textbf{a$_4$} & \textbf{a$_5$} & \textbf{a$_6$} \\
\midrule
THE MAJORITY WILL NEVER allow the Mentally Ill Globalists to turn the world into a SJW and Radical Islam SAFE SPACE \#brexit \#Trump2016 & 0 & 0 & 1 & 1 & 0 & 1 \\
A muslim Mayor of London? What!? This PC Sickness has become a pandemic. England turning into Little Asia. & 0 & 0 & 0 & 1 & 1 & 1 \\
Not all foreign people who wants to go to the uk have bad intentions. Improve your law. The \#Brexit isn’t gonna help your economy. & 0 & 0 & 0 & 0 & 0 & 0 \\
\bottomrule
\end{tabular}
}
\caption{Examples from Brexit dataset with binary Hate labels from all 6 annotators.}
\label{tab:brexit_examples}
\end{table}

\subsection{Demographics of MFSC Annotators} \label{sec:mfrc-diverse}
\begin{figure}[ht]
    \centering
\includegraphics[width=\columnwidth]{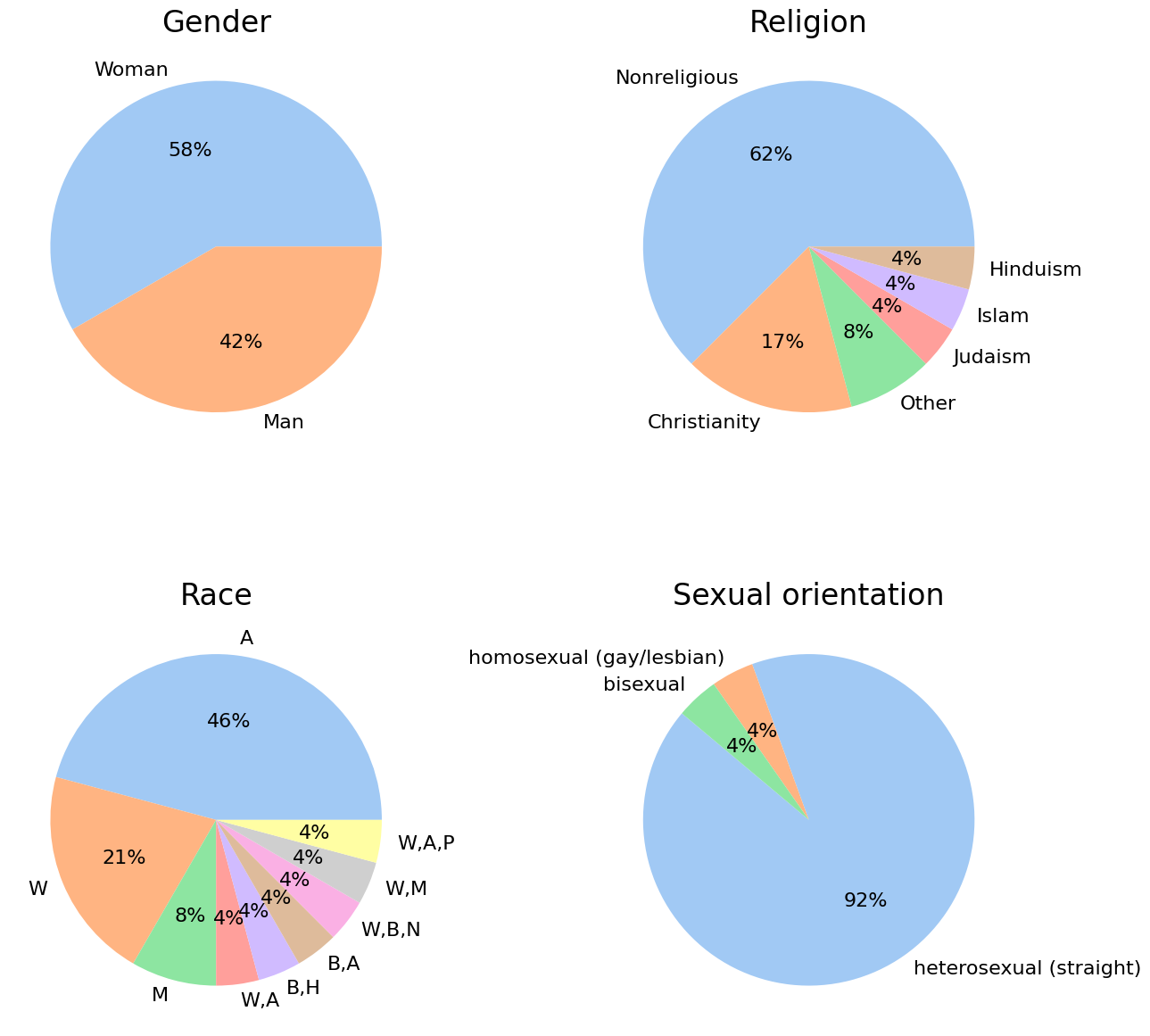}
    \caption{The abbreviations in the pie chart for race  \textit{W} stands for White or European American, \textit{B} stands for Black or African American, \textit{H} stands for Hispanic or Latino/Latinx, \textit{P} stands for Native Hawaiian or Pacific Islander, \textit{A} stands for Asian or Asian American, \textit{M} stands for Middle Eastern or North African.}
    \label{fig:diverse}
    \vspace{-4mm}
\end{figure}

We aimed to diversify the annotators for MFSC dataset across gender, sexual orientation,	religion, and race. Even though our dataset is not balanced across these dimensions, we strived to include representative annotators from a cross-section of the aforementioned demographics. The distribution of the annotators across the mentioned demographics is presented in \autoref{fig:diverse}.

\section{Additional Baseline and Ablation Study} \label{sec:app-ablation}

In the following sections, we conduct the experiments using the Roberta-base model due to its superior performance among the base models in our experiments, as well as its resource efficiency.

\begin{figure}[ht]
\centering
\begin{subfigure}[b]{\columnwidth}
\begin{subfigure}[b]{0.49\columnwidth}
   \centering
   \includegraphics[width=\textwidth, height=3cm]{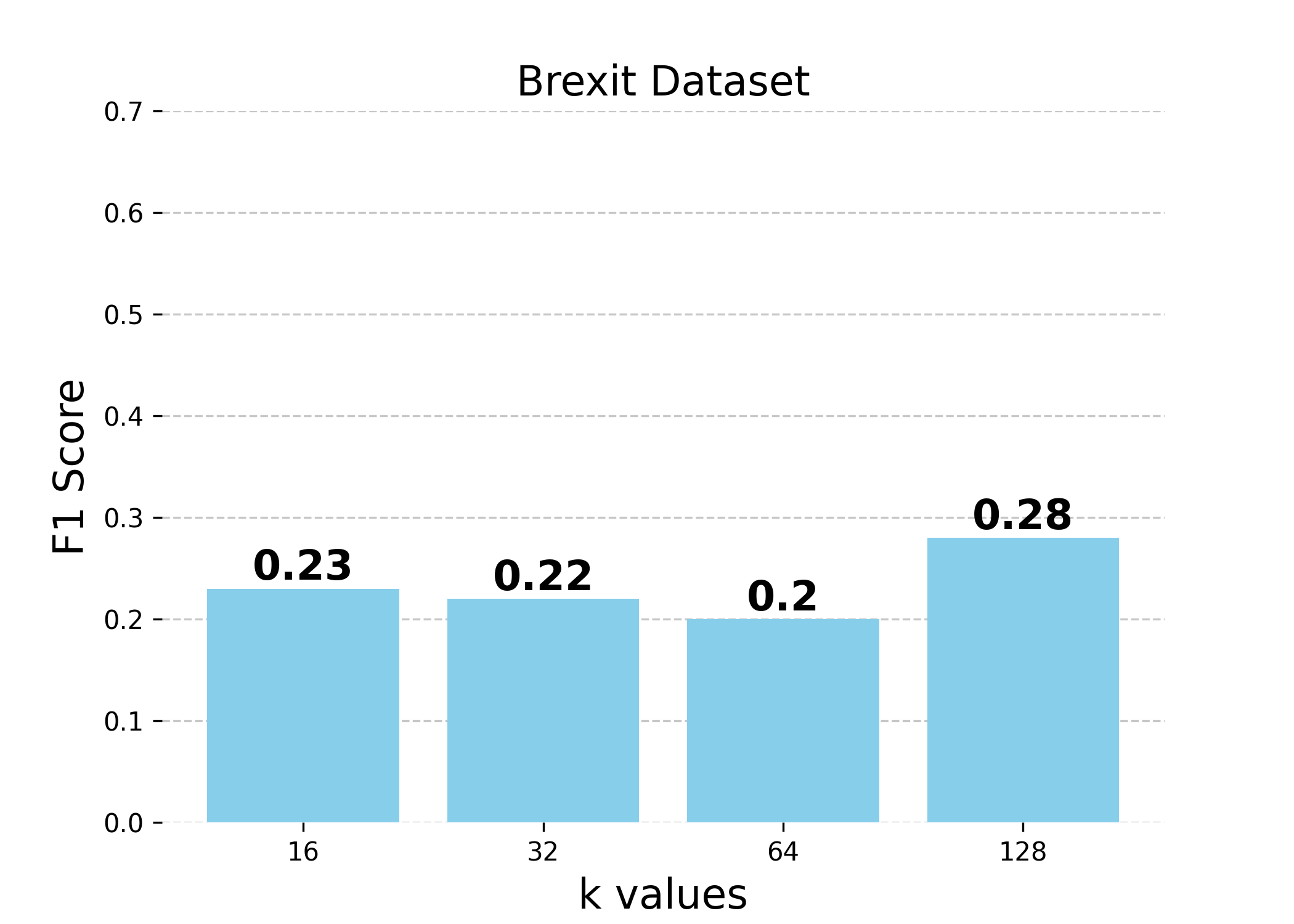}
\end{subfigure}
\begin{subfigure}[b]{0.49\columnwidth}
    \centering
   \includegraphics[width=\textwidth,height=3cm]{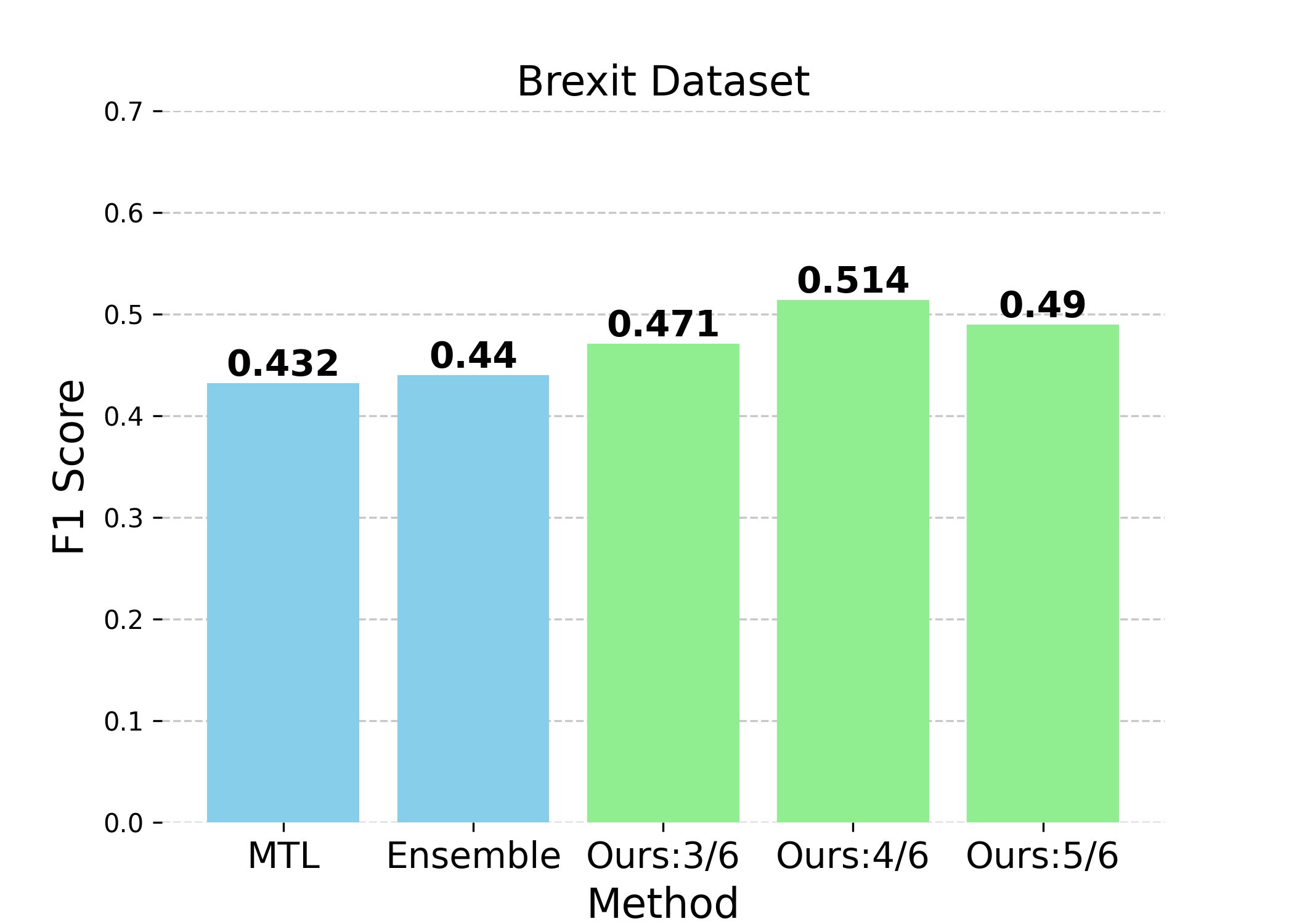}
\end{subfigure}
\caption{Brexit}
\end{subfigure}
\begin{subfigure}[b]{\columnwidth}
\begin{subfigure}[b]{0.49\columnwidth}
   \centering
   \includegraphics[width=\textwidth,height=3cm]{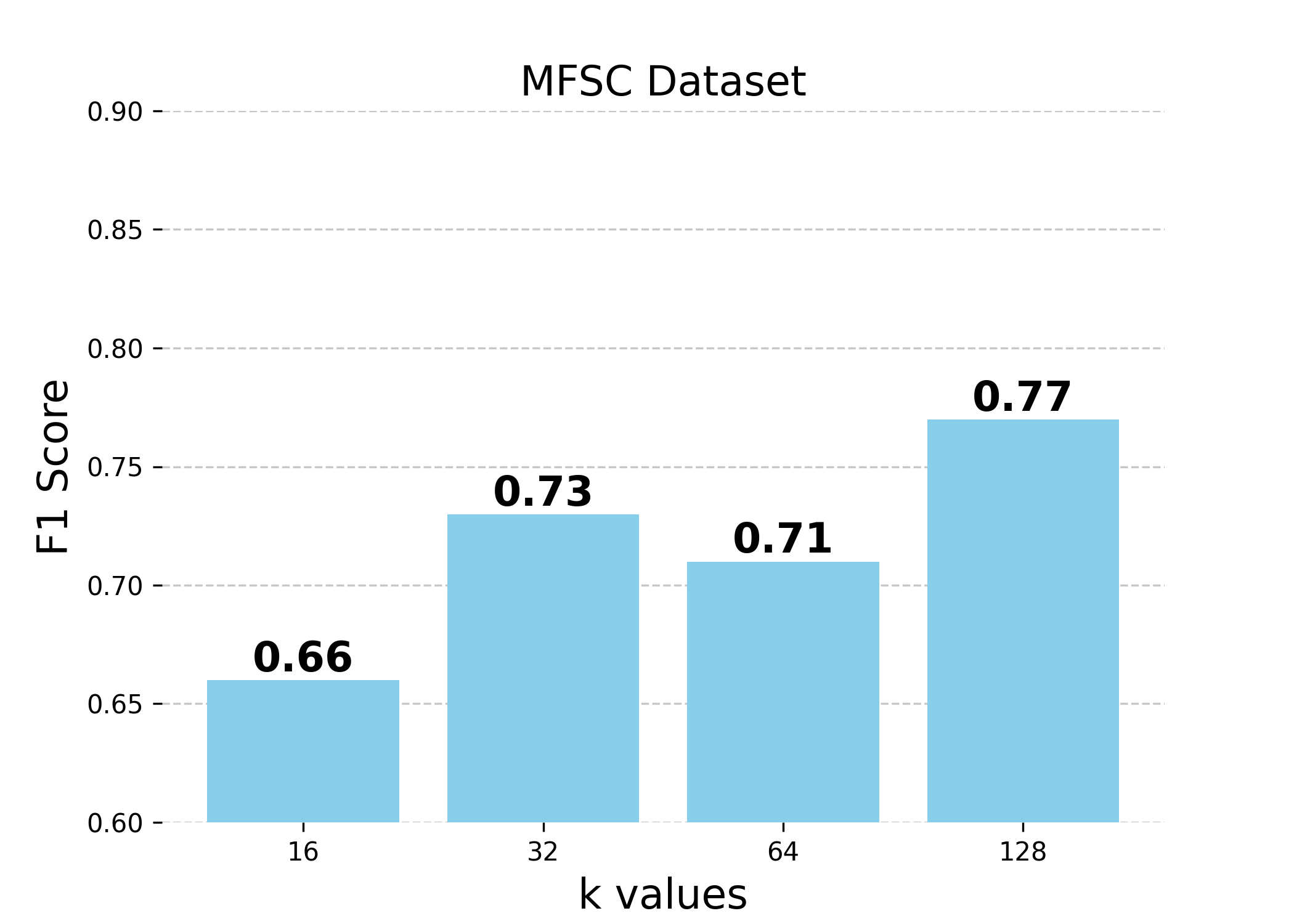}
\end{subfigure}
\begin{subfigure}[b]{0.49\columnwidth}
    \centering
   \includegraphics[width=\textwidth,height=3cm]{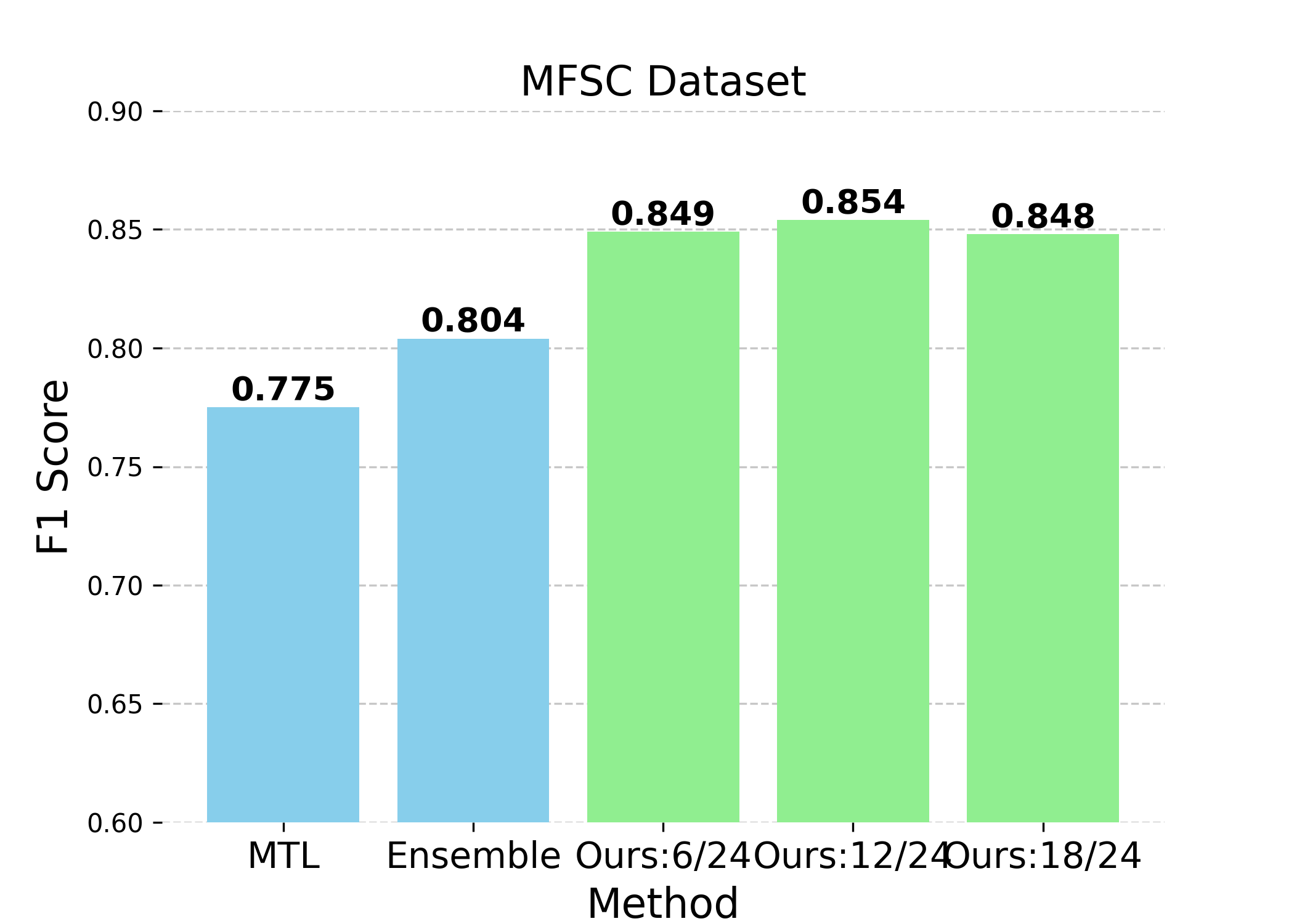}
\end{subfigure}
\caption{MFSC}
\end{subfigure}
\caption{Baseline results in Blue compared to our framework results in green}
\label{fig-app:full_ft}
\end{figure}

First, we conduct an ablation study by omitting the first stage of MTL, effectively reducing the model to few-shot adaptation for each annotator from a pre-trained model. The resulting $F_1$ scores are shown in \autoref{fig-app:full_ft}. When comparing these scores to our complete framework in \autoref{tab_app:overall_res}, we observe that our framework consistently outperforms the second stage alone in all few-shot scenarios. For instance, in random few-shot sampling for $k=16$, our model achieves a 23\% gain in Brexit and a 12\% gain in MFSC compared to this ablation model. This highlights the critical role of the first stage of MTL in the success of our framework.


In the second ablation study, we omit the second stage, few-shot sample selection, from our framework. In other words, in the second stage, we use all annotated samples for each annotator instead of selecting only a few samples. Note that this is equivalent to using 100\% of the budget and serves as an upper bound to the performance achieved with an ideal sampling function. 

Additionally, we present a new baseline where a separate model is trained for each annotator using 100\% of their respective data. Following the naming convention used by \citet{davani2022dealing}, we refer to this baseline as “Ensemble” to ensure consistency with previous work in this field.The Ensemble baseline involves fine-tuning the model directly for each annotator, calculating individual annotator $F_1$  scores, and reporting the average $F_1$  score across annotators. Hyperparameters and epoch numbers for training are consistent with those mentioned for the MTL model in Section \ref{implementation}.  \autoref{fig-app:full_ft} presents a comparison of 3 different strategies, using 100\% of the budget (MTL, Ensemble, and ours). On the Brexit Dataset (top) our framework has as much as 7.4\% performance gain compared to the Ensemble baseline (when using $\frac{4}{6}$ annotators in MTL), and for the MFSC dataset our framework has as large as 5\% gain compared to Ensemble baseline (when using $\frac{12}{24}$ of annotators in MTL). These results show that even considering the 100\% budget, our framework outperforms both baselines, demonstrating the benefit of our two-stage design. \blue{Interestingly, the Ensemble model outperforms MTL for these datasets, contrary to previous research findings comparing these two methods.}






\section{Additional Tasks}
\subsection{MFSC (Care label)} \label{app-sec-care}
We evaluate our framework on an additional binary label of \textit{Care} moral concern from our MFSC dataset. This moral concern is defined as \textit{"Care/Harm: Intuitions about avoiding emotional and physical damage or harm to another individual. It underlies virtues of kindness, gentleness, and nurturing, and vices of meanness, violence, and abuse."} \citep{trager2022moral}. \autoref{tab-app:care_overall} presents the results for this task. Our framework outperforms the baseline MTL approach with $25\%$ and $50\%$ of the annotation budget. Notably, with only $25\%$ of the budget, our framework has a $1.4\%$ gain in $F_1$ score compared to MTL with 100\% budget.
The experiments were conducted with the same hyper-parameter tuning described in Section \ref{implementation}.

\renewcommand{\arraystretch}{1.3}
\setlength{\tabcolsep}{2pt}
\begin{table}[h!]\small
\begin{tabular}{ll|cccc}
\toprule  \multicolumn{2}{c|}{\multirow{2}{*}{$metric = F_1^{Overall} \uparrow$}}  &  \multicolumn{4}{c}{MFSC (Care)} \\
  \multicolumn{2}{c|}{}  & $25\%$ & $50\%$  & $75\%$  &  $100\%$ \\

\toprule
 \multicolumn{2}{c|}{\scriptsize{\wideunderline[7em]{$X\% \times \lvert D_{a_i} \rvert$ }}}
  & \scriptsize{$25\%  \lvert D_{a_i} \rvert$} & \scriptsize{$50\%  \lvert D_{a_i} \rvert$} & \scriptsize{$75\%  \lvert D_{a_i} \rvert$} & \scriptsize{$\lvert D_{a_i} \rvert$} \\[0.1cm]

\multicolumn{2}{c|}{MTL} &  0.474 & 0.476 & 0.49 & 0.469 \\ \toprule

\multicolumn{2}{c|}{\scriptsize{\wideunderline[7em]{$X\% \times \lvert \mathcal{A} \rvert$ }}} 
    & \scriptsize{$50\% \lvert \mathcal{A} \rvert$} & \scriptsize{$66\% \lvert \mathcal{A} \rvert$} & \scriptsize{$83\% \lvert \mathcal{A} \rvert$} & \cellcolor{lightgray} \\ [0.1cm]
\multirow[c]{4}{*}{$k=16$}  & $\mathcal{S}_{bal}$ & 0.462 & 0.471 & 0.485 & \cellcolor{lightgray}\\
 & $\mathcal{S}_{dis}$ & 0.46 & 0.467 & 0.485 & \cellcolor{lightgray} \\
 & $\mathcal{S}_{mv}$ & 0.476 & 0.473 & \textbf{0.49} & \cellcolor{lightgray}\\
 & $\mathcal{S}_{rand}$ & 0.469 & 0.468 & 0.482 & \cellcolor{lightgray}\\
\cline{1-6}
\multirow[c]{4}{*}{$k=32$} & $\mathcal{S}_{bal}$ & 0.467 & 0.477 & 0.487 & \cellcolor{lightgray} \\
 & $\mathcal{S}_{dis}$ & 0.463 & 0.463 & 0.483 & \cellcolor{lightgray}\\
 & $\mathcal{S}_{mv}$ & 0.475 & 0.475 & 0.488 & \cellcolor{lightgray}\\
 & $\mathcal{S}_{rand}$ & 0.47 & 0.468 & 0.484 & \cellcolor{lightgray}\\
\cline{1-6}
\multirow[c]{4}{*}{$k=64$} & $\mathcal{S}_{bal}$ & 0.47 & 0.475 & 0.486 & \cellcolor{lightgray}\\
 & $\mathcal{S}_{dis}$ & 0.467 & 0.471 & 0.478 & \cellcolor{lightgray}\\
 & $\mathcal{S}_{mv}$ & 0.479 & 0.48 & 0.487 & \cellcolor{lightgray}\\
 & $\mathcal{S}_{rand}$ & 0.472 & 0.477 & \textbf{0.49} & \cellcolor{lightgray}\\
\cline{1-6}
\multirow[c]{4}{*}{$k=128$}& $\mathcal{S}_{bal}$ & 0.473 & 0.477 & 0.488 & \cellcolor{lightgray}\\
 & $\mathcal{S}_{dis}$ & 0.474 & 0.474 & 0.481 & \cellcolor{lightgray}\\
 & $\mathcal{S}_{mv}$ & 0.477 & \textbf{0.482} & 0.488 & \cellcolor{lightgray}\\
 & $\mathcal{S}_{rand}$ & \textbf{0.483} & 0.481 & 0.487 & \cellcolor{lightgray}\\
\cline{1-6}
\bottomrule
\end{tabular}
\caption{Overall $F_1$ scores on MFSC dataset, \textit{Care} label, with varying annotation budgets $(\%B)$.}
\label{tab-app:care_overall}
\vspace{-6mm}
\end{table}

\subsection{GHC (Hate label)} \label{sec:ghc}
To ensure the generalizability of our framework, we evaluate it on a larger dataset with an imbalanced number of annotations among annotators. We conducted the experiments using the RoBERTa-Base model due to its superior performance among the base models in our experiments, as well as its resource efficiency.

\noindent\textbf{Gab Hate Corpus (GHC)} consists of 27,665 posts from the social network service gab.ai, each annotated by a minimum of three trained annotators, and 18 total annotators. It is coded for hate-based rhetoric and has labels of “assaults on human dignity” or “calls for violence”. 
\begin{figure}[ht]
    \centering
    \includegraphics[width=0.8\columnwidth]{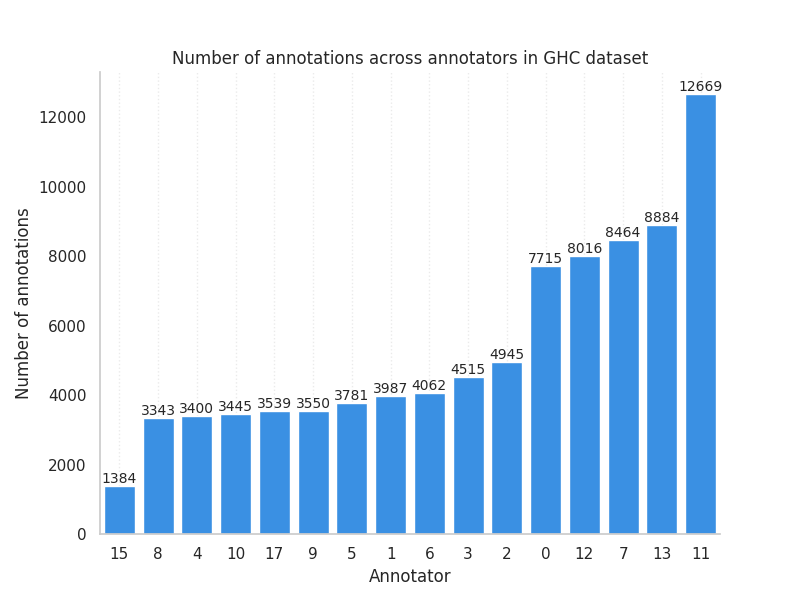}
    \caption{The number of annotated instances by each annotator in GHC dataset}
    \label{fig:ghc_num_annot}
\end{figure}
The annotators with less than 1000 annotations were filtered out resulting in 16 annotators. \autoref{fig:ghc_num_annot} shows the number of annotated instances by each annotator.

\noindent\textbf{Experiments:} We replicate the experiment described in Section \ref{sec:exp-setup} with the same implementation details outlined in Section \ref{implementation}. We employ varying budgets of $25\%$, $50\%$, and $75\%$, using the two best-performing sampling methods identified in our experiments ($\mathcal{S}_{bal}$ and $\mathcal{S}_{rand}$), and compare the results to the MTL baseline. The overall results are presented in \autoref{tab-app:ghc_overall}. It is evident that our framework consistently outperforms MTL across all numbers of shots, sampling methods, and budget variations. Specifically, with $25\%$ of the budget, our model achieves a gain of $1.6\%$ with $k=64$ and $\mathcal{S}_{rand}$, and with $75\%$ of the budget, our model performs the best, achieving a gain of $2\%$.
\renewcommand{\arraystretch}{1.4}
\setlength{\tabcolsep}{3pt}
\begin{table}[h!]\small
\begin{tabular}{ll|cccc}
\toprule  \multicolumn{2}{c|}{\multirow{2}{*}{$ F_1^{Overall} \uparrow$}}  &  \multicolumn{4}{c}{GHC} \\
  \multicolumn{2}{c|}{}  & $25\%$ & $50\%$  & $75\%$  &  $100\%$ \\
\toprule
 \multicolumn{2}{c|}{\scriptsize{\wideunderline[5em]{$X\% \times \lvert D_{a_i} \rvert$ }}}
  & \scriptsize{$25\%  \lvert D_{a_i} \rvert$} & \scriptsize{$50\%  \lvert D_{a_i} \rvert$} & \scriptsize{$75\%  \lvert D_{a_i} \rvert$} & \scriptsize{$\lvert D_{a_i} \rvert$} \\[0.1cm]

\multicolumn{2}{c|}{MTL} &  .417\textsubscript{(.004)} & .433\textsubscript{(.007)} & .442\textsubscript{(.013)} & .451\textsubscript{(.006)} \\ \toprule
\multicolumn{2}{c|}{\scriptsize{\wideunderline[7em]{$X\% \times \lvert \mathcal{A} \rvert$ }}} 
    & \scriptsize{$50\% \lvert \mathcal{A} \rvert$} & \scriptsize{$66\% \lvert \mathcal{A} \rvert$} & \scriptsize{$83\% \lvert \mathcal{A} \rvert$} & \cellcolor{lightgray} \\ [0.1cm]
\multirow[c]{2}{*}{$k=16$}  & $\mathcal{S}_{bal}$ & .45\textsubscript{(.004)} & .46\textsubscript{(.002)} & .464\textsubscript{(.003)} & \cellcolor{lightgray} \\
 & $\mathcal{S}_{rand}$ & .455\textsubscript{(.008)} & .469\textsubscript{(.005)} & .468\textsubscript{(.003)}   & \cellcolor{lightgray}\\
\cline{1-6}
\multirow[c]{2}{*}{$k=32$} & $\mathcal{S}_{bal}$ & .456\textsubscript{(.002)} & .459\textsubscript{(.001)} & .464\textsubscript{(.003)}  & \cellcolor{lightgray} \\
 & $\mathcal{S}_{rand}$ & .461\textsubscript{(.003)} & .472\textsubscript{(.001)} & .468\textsubscript{(.001)}  & \cellcolor{lightgray}\\
\cline{1-6}
\multirow[c]{2}{*}{$k=64$} & $\mathcal{S}_{bal}$ & .458\textsubscript{(.004)} & .461\textsubscript{(.002)} & .466\textsubscript{(.003)} & \cellcolor{lightgray} \\
 & $\mathcal{S}_{rand}$ & \textbf{.467}\textsubscript{(.003)} & .474\textsubscript{(.002)} & .468\textsubscript{(.002)} & \cellcolor{lightgray} \\
\cline{1-6}
\multirow[c]{2}{*}{$k=128$}& $\mathcal{S}_{bal}$ & .466\textsubscript{(.001)} & .466\textsubscript{(0)} & .466\textsubscript{(.003)} & \cellcolor{lightgray} \\
 & $\mathcal{S}_{rand}$ & .463\textsubscript{(.007)} & \textbf{.475}\textsubscript{(.003)} & \textbf{.47}\textsubscript{(.001)} & \cellcolor{lightgray} \\
\cline{1-6}
\bottomrule
\end{tabular}
\caption{Overall $F_1$ scores on GHC dataset, \textit{Hate} label, with varying annotation budgets $(\%B)$.}
\label{tab-app:ghc_overall}
\end{table}

\begin{figure}[ht!]
    \centering
    \includegraphics[width=0.8\columnwidth]{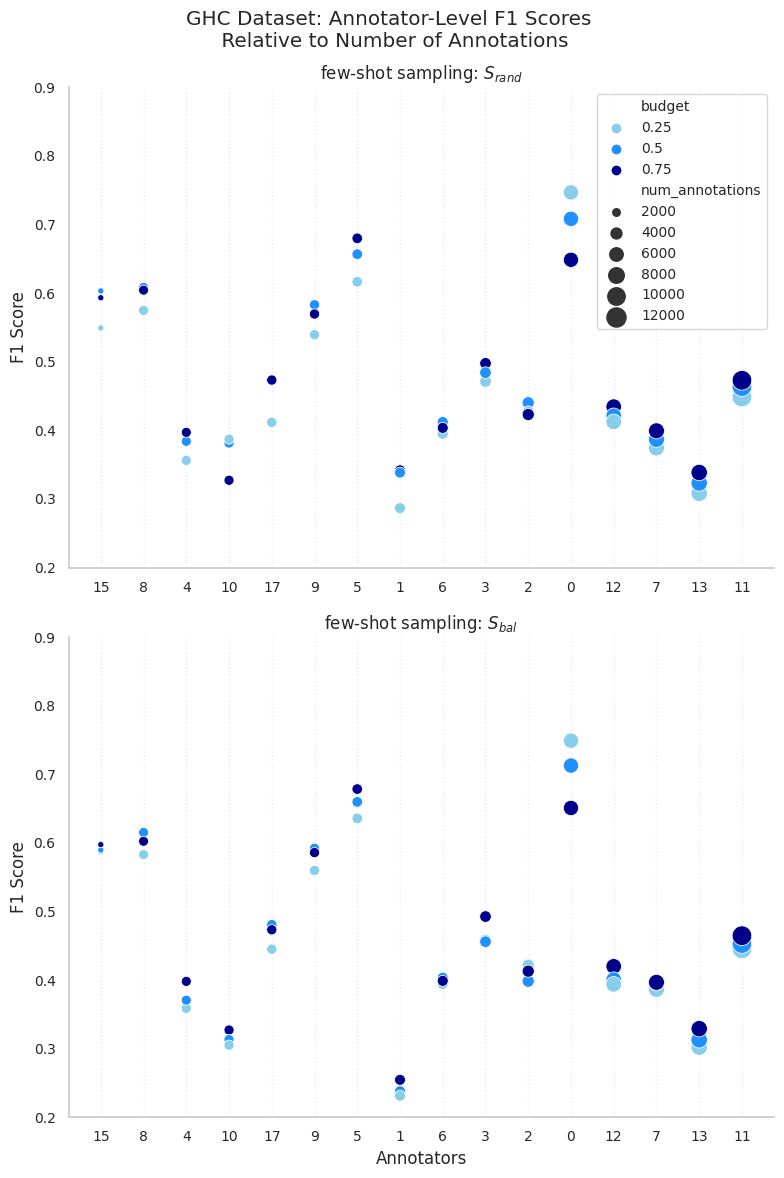}
    \caption{$F_1$ scores of annotators as the number of annotations increases}
    \label{fig:ghc_corr}
\end{figure}

\noindent\textbf{Impact of the Imbalanced Number of Annotations on Performance}
Results on the GHC dataset indicate a consistent and significant advantage of our framework, even when applied to larger datasets with imbalanced numbers of annotations across annotators. To further investigate the impact of varying numbers of annotations across annotators on the performance of our framework, we conducted a correlation analysis between each annotator's performance and their number of annotations. The results revealed no statistically significant correlation between the number of annotations and the overall $F_1$ score of an annotator, as indicated by the correlation coefficients for $\mathcal{S}_{rand}$ $(r=-0.17, p=0.25)$ and $\mathcal{S}_{bal}$ $(r=-0.14, p=0.32)$. The plots in \autoref{fig:ghc_corr} illustrate the annotator-level $F_1$ scores as the number of annotations of the annotators increases.

\section{Additional Details and Results} \label{app-result_details}

Here, we present the results of our framework for all values of $k$, with the mean and standard deviations reported for three seeds for the RoBERTa models. Tables \ref{tab_app:overall_res}, \ref{roberta-large_overall_f1}, and \ref{llama3_overall_f1} show the results for RoBERTa-Base, RoBERTa-Large, and Llama-3, respectively. The best values are highlighted in \textbf{bold}. As evident, our framework outperforms the baseline across all three models.

\subsection{Implementation Details} \label{sec:app-imp_details}
For the RoBERTa models, hyperparameter tuning was conducted for each MTL model with learning rates of $[3\text{e-06}, 5\text{e-05}, 1\text{e-06}, 2\text{e-05}]$ and weight decays of $[0, 0.01]$. For the Llama model, hyperparameter tuning included learning rate, weight decay, LoRA alpha, LoRA rank, and LoRA dropout for one MTL model, and these parameters were used across all models. The best configuration and other parameters for training Llama-3 with LoRA are shown in \autoref{tab:hyperparameters}.


\renewcommand{\arraystretch}{1.2}
\setlength{\tabcolsep}{5pt}
\begin{table}[ht]
    \centering
    \small
  
    \begin{tabular}{lcc}
        \toprule
        \textbf{Hyperparameter} & \textbf{Brexit} & \textbf{MFSC} \\
        \midrule
        Train Batch Size & 16 & 4 \\
        Eval Steps & 50 & 100 \\
        Max Length & 512 & 512 \\
        Learning Rate & 1.2e-04 & 5e-05 \\
        Epochs & 10 & 2 \\
        Weight Decay & 0.01 & 0.01 \\
        LoRA r & 8 & 4 \\
        LoRA Alpha & 32 & 16 \\
        LoRA Dropout & 0.003 & 0.1 \\
        \bottomrule
    \end{tabular}
    \caption{Hyperparameters of the Lora Llama-3 model trained for Brexit and MFSC Datasets}
    \label{tab:hyperparameters}
\end{table}

\setlength{\tabcolsep}{4pt}
\renewcommand{\arraystretch}{1.2}
\begin{table*}[ht]\small
\centering
\begin{tabular}{ll|cccc|cccc}
\toprule  \multicolumn{2}{c|}{\multirow{2}{*}{$metric = F_1^{overall} \uparrow$}} & \multicolumn{4}{c|}{Brexit} & \multicolumn{4}{c}{\mf} \\
\multicolumn{2}{c|}{} & $50\%$ & $66\%$ & $83\%$ & $100\%$ & $25\%$ & $50\%$ & $75\%$ & $100\%$ \\
\toprule
 \multicolumn{2}{c|}{\scriptsize{\wideunderline[7em]{$X\% \times \lvert D_{a_i} \rvert$ }}}
  & \scriptsize{$50\%  \lvert D_{a_i} \rvert$} & \scriptsize{$66\%  \lvert D_{a_i} \rvert$} & \scriptsize{$83\%  \lvert D_{a_i} \rvert$} & \scriptsize{$\lvert D_{a_i} \rvert$} & \scriptsize{$25\% \lvert D_{a_i} \rvert$} & \scriptsize{$50\%  \lvert D_{a_i} \rvert$} & \scriptsize{$ 75\% \lvert D_{a_i} \rvert$} & \scriptsize{$\lvert D_{a_i} \rvert$} \\ [0.1cm]

\multicolumn{2}{c|}{MTL} & 0.417\textsubscript{(0.049)} & 0.449\textsubscript{(0.027)} & 0.418\textsubscript{(0.018)} & 0.431\textsubscript{(0.014)} & 0.763\textsubscript{(0.016)} & 0.773\textsubscript{(0.011)} & 0.772\textsubscript{(0.015)} & 0.776\textsubscript{(0.004)} \\ [0.1cm]
\toprule

 \multicolumn{2}{c|}{\scriptsize{\wideunderline[7em]{$X\% \times \lvert \mathcal{A} \rvert$ }}} 
    & \scriptsize{$50\% \lvert \mathcal{A} \rvert$} & \scriptsize{$66\% \lvert \mathcal{A} \rvert$} & \scriptsize{$83\% \lvert \mathcal{A} \rvert$} & \cellcolor{lightgray} 
    & \scriptsize{$ 25\% \lvert \mathcal{A}  \rvert$} & \scriptsize{$ 50\% \lvert \mathcal{A}  \rvert$} & \scriptsize{$ 75\% \lvert \mathcal{A}  \rvert$} & \cellcolor{lightgray}  \\ [0.1cm]

\multirow[c]{4}{*}{$k=16$} 
 & $\mathcal{S}_{bal}$ & 0.443\textsubscript{(0.005)} & 0.454\textsubscript{(0.015)} & 0.457\textsubscript{(0.015)}  & \cellcolor{lightgray} & 0.777\textsubscript{(0.002)} & 0.779\textsubscript{(0.0)} & 0.78\textsubscript{(0.003)} & \cellcolor{lightgray} \\
 & $\mathcal{S}_{dis}$ & 0.421\textsubscript{(0.01)} & 0.44\textsubscript{(0.011)} & 0.457\textsubscript{(0.008)}  & \cellcolor{lightgray} & 0.784\textsubscript{(0.01)} & 0.787\textsubscript{(0.004)} & 0.782\textsubscript{(0.005)}  & \cellcolor{lightgray}\\
 & $\mathcal{S}_{mv}$ & 0.426\textsubscript{(0.008)} & 0.441\textsubscript{(0.019)} & 0.455\textsubscript{(0.019)}  & \cellcolor{lightgray} & 0.789\textsubscript{(0.004)} & 0.786\textsubscript{(0.004)} & 0.783\textsubscript{(0.006)}  & \cellcolor{lightgray} \\
 & $\mathcal{S}_{rand}$ & 0.422\textsubscript{(0.012)} & 0.44\textsubscript{(0.025)} & 0.455\textsubscript{(0.007)}  & \cellcolor{lightgray} & 0.795\textsubscript{(0.009)} & 0.79\textsubscript{(0.006)} & 0.785\textsubscript{(0.005)}  & \cellcolor{lightgray} \\
\cline{1-10}
\multirow[c]{4}{*}{$k=32$} 
 & $\mathcal{S}_{bal}$ & 0.449\textsubscript{(0.008)} & 0.458\textsubscript{(0.009)} & 0.458\textsubscript{(0.008)}  & \cellcolor{lightgray} & 0.779\textsubscript{(0.002)} & 0.78\textsubscript{(0.001)} & 0.78\textsubscript{(0.003)}  & \cellcolor{lightgray} \\
 & $\mathcal{S}_{dis}$ & 0.423\textsubscript{(0.008)} & 0.44\textsubscript{(0.015)} & 0.457\textsubscript{(0.016)}  & \cellcolor{lightgray} & 0.786\textsubscript{(0.01)} & 0.788\textsubscript{(0.004)} & 0.783\textsubscript{(0.005)}  & \cellcolor{lightgray} \\
 & $\mathcal{S}_{mv}$ & 0.424\textsubscript{(0.017)} & 0.444\textsubscript{(0.02)} & 0.458\textsubscript{(0.011)}  & \cellcolor{lightgray} & 0.791\textsubscript{(0.004)} & 0.787\textsubscript{(0.003)} & 0.783\textsubscript{(0.007)}  & \cellcolor{lightgray}\\
 & $\mathcal{S}_{rand}$ & 0.428\textsubscript{(0.006)} & 0.447\textsubscript{(0.019)} & 0.452\textsubscript{(0.016)}  & \cellcolor{lightgray} & 0.795\textsubscript{(0.01)} & \textbf{0.791}\textsubscript{(0.006)} & 0.785\textsubscript{(0.004)}  & \cellcolor{lightgray}\\
\cline{1-10}
\multirow[c]{4}{*}{$k=64$}
 & $\mathcal{S}_{bal}$ & 0.453\textsubscript{(0.003)} & 0.458\textsubscript{(0.016)} & 0.459\textsubscript{(0.011)}  & \cellcolor{lightgray} & 0.78\textsubscript{(0.003)} & 0.781\textsubscript{(0.003)} & 0.781\textsubscript{(0.004)}  & \cellcolor{lightgray} \\
 & $\mathcal{S}_{dis}$ & 0.436\textsubscript{(0.01)} & 0.455\textsubscript{(0.016)} & 0.468\textsubscript{(0.01)}  & \cellcolor{lightgray} & 0.787\textsubscript{(0.01)} & 0.789\textsubscript{(0.004)} & 0.783\textsubscript{(0.005)}  & \cellcolor{lightgray}\\
 & $\mathcal{S}_{mv}$ & 0.427\textsubscript{(0.007)} & 0.439\textsubscript{(0.026)} & 0.459\textsubscript{(0.013)}  & \cellcolor{lightgray} & 0.791\textsubscript{(0.005)} & 0.788\textsubscript{(0.003)} & 0.784\textsubscript{(0.007)}  & \cellcolor{lightgray} \\
 & $\mathcal{S}_{rand}$ & 0.433\textsubscript{(0.012)} & 0.451\textsubscript{(0.015)} & 0.456\textsubscript{(0.013)}  & \cellcolor{lightgray} & 0.797\textsubscript{(0.009)} & 0.791\textsubscript{(0.006)} & 0.785\textsubscript{(0.004)}  & \cellcolor{lightgray}\\
\cline{1-10}
\multirow[c]{4}{*}{$k=128$} 
 & $\mathcal{S}_{bal}$ & \textbf{0.471}\textsubscript{(0.002)} & \textbf{0.474}\textsubscript{(0.018)} & \textbf{0.468}\textsubscript{(0.014)}  & \cellcolor{lightgray} & 0.781\textsubscript{(0.002)} & 0.781\textsubscript{(0.002)} & 0.782\textsubscript{(0.003)}  & \cellcolor{lightgray} \\
 & $\mathcal{S}_{dis}$ & 0.45\textsubscript{(0.008)} & 0.461\textsubscript{(0.019)} & 0.466\textsubscript{(0.016)}  & \cellcolor{lightgray} & 0.788\textsubscript{(0.009)} & 0.789\textsubscript{(0.003)} & 0.783\textsubscript{(0.005)}  & \cellcolor{lightgray} \\
 & $\mathcal{S}_{mv}$ & 0.434\textsubscript{(0.015)} & 0.445\textsubscript{(0.022)} & 0.458\textsubscript{(0.016)}  & \cellcolor{lightgray}  & 0.793\textsubscript{(0.005)} & 0.788\textsubscript{(0.004)} & 0.784\textsubscript{(0.007)}  & \cellcolor{lightgray} \\
 & $\mathcal{S}_{rand}$ & 0.439\textsubscript{(0.015)} & 0.457\textsubscript{(0.012)} & 0.455\textsubscript{(0.011)}  & \cellcolor{lightgray} & \textbf{0.798}\textsubscript{(0.008)} & 0.791\textsubscript{(0.005)} & \textbf{0.786}\textsubscript{(0.004)}  & \cellcolor{lightgray} \\
\cline{1-10}
\bottomrule
\end{tabular}
\caption{\textbf{RoBERTa-Base} Overall $F_1$ results on Brexit  and \mf datasets for different budgets of annotation $(B)$, with various few shot sampling strategies; mean and standard deviation calculated over repeated runs.}
\label{tab_app:overall_res}
\end{table*}


\begin{table*}[h]\small

\centering
\begin{tabular}{ll|cccc|cccc}
\toprule  \multicolumn{2}{c|}{\multirow{2}{*}{$metric = F_1^{overall} \uparrow$}} & \multicolumn{4}{c|}{Brexit} & \multicolumn{4}{c}{\mf} \\
\multicolumn{2}{c|}{} & $50\%$ & $66\%$ & $83\%$ & $100\%$ & $25\%$ & $50\%$ & $75\%$ & $100\%$ \\
\toprule
 \multicolumn{2}{c|}{\scriptsize{\wideunderline[7em]{$X\% \times \lvert D_{a_i} \rvert$ }}}
  & \scriptsize{$50\%  \lvert D_{a_i} \rvert$} & \scriptsize{$66\%  \lvert D_{a_i} \rvert$} & \scriptsize{$83\%  \lvert D_{a_i} \rvert$} & \scriptsize{$\lvert D_{a_i} \rvert$} & \scriptsize{$25\% \lvert D_{a_i} \rvert$} & \scriptsize{$50\%  \lvert D_{a_i} \rvert$} & \scriptsize{$ 75\% \lvert D_{a_i} \rvert$} & \scriptsize{$\lvert D_{a_i} \rvert$} \\ [0.1cm]

\multicolumn{2}{c|}{MTL} & 0.366\textsubscript{(0.123)} & 0.476\textsubscript{(0.026)} & 0.497\textsubscript{(0.012)} & 0.475\textsubscript{(0.012)} & 0.773\textsubscript{(0.002)} & 0.768\textsubscript{(0.007)} & 0.772\textsubscript{(0.004)} & 0.771\textsubscript{(0.004)} \\ [0.1cm]

\toprule

\multicolumn{2}{c|}{\scriptsize{\wideunderline[7em]{$X\% \times \lvert \mathcal{A} \rvert$ }}} 
    & \scriptsize{$50\% \lvert \mathcal{A} \rvert$} & \scriptsize{$66\% \lvert \mathcal{A} \rvert$} & \scriptsize{$83\% \lvert \mathcal{A} \rvert$} & \cellcolor{lightgray} 
    & \scriptsize{$ 25\% \lvert \mathcal{A}  \rvert$} & \scriptsize{$ 50\% \lvert \mathcal{A}  \rvert$} & \scriptsize{$ 75\% \lvert \mathcal{A}  \rvert$} & \cellcolor{lightgray}  \\ [0.1cm]

\multirow[c]{4}{*}{$k=16$} & $\mathcal{S}_{bal}$ & 0.48\textsubscript{(0.003)} & 0.476\textsubscript{(0.008)} & 0.484\textsubscript{(0.01)} & \cellcolor{lightgray} & 0.778\textsubscript{(0.003)} & 0.779\textsubscript{(0.003)} & 0.776\textsubscript{(0.002)} & \cellcolor{lightgray} \\
 & $\mathcal{S}_{dis}$ & 0.475\textsubscript{(0.01)} & 0.471\textsubscript{(0.003)} & 0.487\textsubscript{(0.012)} & \cellcolor{lightgray} & 0.787\textsubscript{(0.001)} & 0.787\textsubscript{(0.001)} & 0.779\textsubscript{(0.002)} & \cellcolor{lightgray} \\
 & $\mathcal{S}_{mv}$ & 0.463\textsubscript{(0.027)} & 0.47\textsubscript{(0.003)} & 0.486\textsubscript{(0.018)} & \cellcolor{lightgray} & 0.786\textsubscript{(0.001)} & 0.786\textsubscript{(0.002)} & 0.779\textsubscript{(0.002)} & \cellcolor{lightgray} \\
 & $\mathcal{S}_{rand}$ & 0.477\textsubscript{(0.019)} & 0.474\textsubscript{(0.006)} & 0.486\textsubscript{(0.02)} & \cellcolor{lightgray} & 0.787\textsubscript{(0.0)} & 0.786\textsubscript{(0.001)} & 0.779\textsubscript{(0.003)} & \cellcolor{lightgray} \\
\cline{1-10}
\multirow[c]{4}{*}{$k=32$} & $\mathcal{S}_{bal}$ & 0.486\textsubscript{(0.003)} & 0.479\textsubscript{(0.005)} & 0.486\textsubscript{(0.011)} & \cellcolor{lightgray} & 0.777\textsubscript{(0.003)} & 0.778\textsubscript{(0.002)} & 0.776\textsubscript{(0.002)} & \cellcolor{lightgray} \\
 & $\mathcal{S}_{dis}$ & 0.475\textsubscript{(0.006)} & 0.477\textsubscript{(0.009)} & 0.484\textsubscript{(0.014)} & \cellcolor{lightgray} & 0.789\textsubscript{(0.002)} & 0.787\textsubscript{(0.001)} & 0.78\textsubscript{(0.002)} & \cellcolor{lightgray} \\
 & $\mathcal{S}_{mv}$ & 0.475\textsubscript{(0.026)} & 0.471\textsubscript{(0.005)} & 0.487\textsubscript{(0.013)} & \cellcolor{lightgray} & 0.788\textsubscript{(0.001)} & 0.786\textsubscript{(0.001)} & 0.779\textsubscript{(0.002)} & \cellcolor{lightgray} \\
 & $\mathcal{S}_{rand}$ & 0.485\textsubscript{(0.004)} & 0.473\textsubscript{(0.004)} & 0.485\textsubscript{(0.016)} & \cellcolor{lightgray} & 0.787\textsubscript{(0.0)} & 0.787\textsubscript{(0.001)} & 0.779\textsubscript{(0.003)} & \cellcolor{lightgray} \\
\cline{1-10}
\multirow[c]{4}{*}{$k=64$} & $\mathcal{S}_{bal}$ & 0.492\textsubscript{(0.006)} & 0.48\textsubscript{(0.002)} & 0.487\textsubscript{(0.009)} & \cellcolor{lightgray} & 0.775\textsubscript{(0.004)} & 0.779\textsubscript{(0.002)} & 0.777\textsubscript{(0.001)} & \cellcolor{lightgray} \\
 & $\mathcal{S}_{dis}$ & 0.501\textsubscript{(0.003)} & 0.479\textsubscript{(0.005)} & \textbf{0.499}\textsubscript{(0.011)} & \cellcolor{lightgray} & 0.79\textsubscript{(0.001)} & 0.788\textsubscript{(0.001)} & 0.78\textsubscript{(0.002)} & \cellcolor{lightgray} \\
 & $\mathcal{S}_{mv}$ & 0.479\textsubscript{(0.025)} & 0.474\textsubscript{(0.004)} & 0.487\textsubscript{(0.013)} & \cellcolor{lightgray} & 0.79\textsubscript{(0.001)} & 0.787\textsubscript{(0.001)} & 0.78\textsubscript{(0.002)} & \cellcolor{lightgray} \\
 & $\mathcal{S}_{rand}$ & 0.49\textsubscript{(0.008)} & 0.479\textsubscript{(0.007)} & 0.486\textsubscript{(0.004)} & \cellcolor{lightgray} & 0.79\textsubscript{(0.002)} & 0.787\textsubscript{(0.001)} & 0.78\textsubscript{(0.003)} & \cellcolor{lightgray} \\
\cline{1-10}
\multirow[c]{4}{*}{$k=128$} & $\mathcal{S}_{bal}$ & \textbf{0.513}\textsubscript{(0.006)} & \textbf{0.492}\textsubscript{(0.008)} & 0.493\textsubscript{(0.005)} & \cellcolor{lightgray} & 0.779\textsubscript{(0.005)} & 0.78\textsubscript{(0.003)} & 0.777\textsubscript{(0.001)} & \cellcolor{lightgray} \\
 & $\mathcal{S}_{dis}$ & 0.506\textsubscript{(0.002)} & 0.486\textsubscript{(0.002)} & 0.497\textsubscript{(0.01)} & \cellcolor{lightgray} & \textbf{0.791}\textsubscript{(0.001)} & \textbf{0.79}\textsubscript{(0.0)} & \textbf{0.781}\textsubscript{(0.002)} & \cellcolor{lightgray} \\
 & $\mathcal{S}_{mv}$ & 0.478\textsubscript{(0.031)} & 0.479\textsubscript{(0.005)} & 0.489\textsubscript{(0.014)} & \cellcolor{lightgray} & \textbf{0.791}\textsubscript{(0.001)} & 0.788\textsubscript{(0.0)} & \textbf{0.781}\textsubscript{(0.002)} & \cellcolor{lightgray} \\
 & $\mathcal{S}_{rand}$ & 0.495\textsubscript{(0.004)} & 0.485\textsubscript{(0.01)} & 0.494\textsubscript{(0.01)} & \cellcolor{lightgray} & \textbf{0.791}\textsubscript{(0.002)} & 0.788\textsubscript{(0.001)} & \textbf{0.781}\textsubscript{(0.003)} & \cellcolor{lightgray} \\
\cline{1-10}
\bottomrule
\end{tabular}
\caption{\textbf{RoBERTa-Large} overall Aggregated $F_1$ results on BREXIT  and MFRC dataset for different $\%B_f$s of annotation, mean and std over 3  runs}
\label{roberta-large_overall_f1}
\end{table*}

\setlength{\tabcolsep}{4pt}
\begin{table*}[h!]\small
\centering
\begin{tabular}{ll|cccc|cccc}
\toprule  \multicolumn{2}{c|}{\multirow{2}{*}{$metric = F_1^{overall} \uparrow$}} & \multicolumn{4}{c|}{Brexit} & \multicolumn{4}{c}{\mf} \\
\multicolumn{2}{c|}{} & $50\%$ & $66\%$ & $83\%$ & $100\%$ & $25\%$ & $50\%$ & $75\%$ & $100\%$ \\
\toprule
 \multicolumn{2}{c|}{\scriptsize{\wideunderline[7em]{$X\% \times \lvert D_{a_i} \rvert$ }}}
  & \scriptsize{$50\%  \lvert D_{a_i} \rvert$} & \scriptsize{$66\%  \lvert D_{a_i} \rvert$} & \scriptsize{$83\%  \lvert D_{a_i} \rvert$} & \scriptsize{$\lvert D_{a_i} \rvert$} & \scriptsize{$25\% \lvert D_{a_i} \rvert$} & \scriptsize{$50\%  \lvert D_{a_i} \rvert$} & \scriptsize{$ 75\% \lvert D_{a_i} \rvert$} & \scriptsize{$\lvert D_{a_i} \rvert$} \\ [0.1cm]

\multicolumn{2}{c|}{MTL} & 0.335 & 0.345 & 0.366 & 0.351  & 0.669 & 0.696 & 0.715 & 0.713 \\ [0.1cm]

\toprule

\multicolumn{2}{c|}{\scriptsize{\wideunderline[7em]{$X\% \times \lvert \mathcal{A} \rvert$ }}} 
    & \scriptsize{$50\% \lvert \mathcal{A} \rvert$} & \scriptsize{$66\% \lvert \mathcal{A} \rvert$} & \scriptsize{$83\% \lvert \mathcal{A} \rvert$} & \cellcolor{lightgray} 
    & \scriptsize{$ 25\% \lvert \mathcal{A}  \rvert$} & \scriptsize{$ 50\% \lvert \mathcal{A}  \rvert$} & \scriptsize{$ 75\% \lvert \mathcal{A}  \rvert$} & \cellcolor{lightgray}  \\ [0.1cm]
\multirow[c]{4}{*}{$k=16$} & $\mathcal{S}_{bal}$ & 0.316 & 0.326 & 0.355 & \cellcolor{lightgray} & 0.637 & 0.648 & 0.692 & \cellcolor{lightgray} \\
 & dis & 0.314 & 0.312 & 0.353 & \cellcolor{lightgray} & 0.628 & 0.678 & 0.683 & \cellcolor{lightgray} \\
 & $\mathcal{S}_{mv}$ & 0.318 & 0.32 & 0.353 & \cellcolor{lightgray} & 0.66 & 0.683 & 0.698 & \cellcolor{lightgray} \\
 & $\mathcal{S}_{rand}$ & 0.294 & 0.32 & 0.35 & \cellcolor{lightgray} & 0.679 & 0.693 & 0.701 & \cellcolor{lightgray} \\
\cline{1-10}
\multirow[c]{4}{*}{$k=32$} & $\mathcal{S}_{bal}$ & 0.337 & 0.338 & 0.363 & \cellcolor{lightgray} & 0.634 & 0.666 & 0.696 & \cellcolor{lightgray} \\
 & dis & 0.318 & 0.323 & 0.353 & \cellcolor{lightgray} & 0.655 & 0.675 & 0.699 & \cellcolor{lightgray} \\
 & $\mathcal{S}_{mv}$ & 0.329 & 0.322 & 0.357 & \cellcolor{lightgray} & 0.681 & 0.691 & 0.7 & \cellcolor{lightgray} \\
 & $\mathcal{S}_{rand}$ & 0.32 & 0.329 & 0.345 & \cellcolor{lightgray} & 0.69 & 0.693 & 0.703 & \cellcolor{lightgray} \\
\cline{1-10}
\multirow[c]{4}{*}{$k=64$} & $\mathcal{S}_{bal}$ & 0.348 & 0.355 & 0.373 & \cellcolor{lightgray} & 0.644 & 0.666 & 0.69 & \cellcolor{lightgray} \\
 & dis & 0.327 & 0.326 & 0.351 & \cellcolor{lightgray} & 0.656 & 0.673 & 0.691 & \cellcolor{lightgray} \\
 & $\mathcal{S}_{mv}$ & 0.359 & 0.339 & 0.365 & \cellcolor{lightgray} & 0.685 & 0.691 & 0.703 & \cellcolor{lightgray} \\
 & $\mathcal{S}_{rand}$ & 0.338 & 0.332 & 0.361 & \cellcolor{lightgray} & 0.703 & 0.706 & 0.705 & \cellcolor{lightgray} \\
\cline{1-10}
\multirow[c]{4}{*}{$k=128$} & $\mathcal{S}_{bal}$ & \textbf{0.384} & \textbf{0.365} & \textbf{0.379} & \cellcolor{lightgray} & 0.675 & 0.685 & 0.704 & \cellcolor{lightgray} \\
 & dis & 0.337 & 0.339 & 0.357 & \cellcolor{lightgray} & 0.664 & 0.688 & 0.698 & \cellcolor{lightgray} \\
 & $\mathcal{S}_{mv}$ & 0.365 & 0.363 & 0.375 & \cellcolor{lightgray} & 0.698 & 0.698 & 0.705 & \cellcolor{lightgray} \\
 & $\mathcal{S}_{rand}$ & 0.339 & 0.343 & 0.365 & \cellcolor{lightgray} & \textbf{0.711} & \textbf{0.713} & \textbf{0.716} & \cellcolor{lightgray} \\
\cline{1-10}
\bottomrule
\end{tabular}
\caption{\textbf{Llama-3} overall Aggregated $F_1$ results on BREXIT  and MFRC dataset for different $\%B_f$s of annotation}
\label{llama3_overall_f1}
\end{table*}
\subsection{Hardware Configuration}
The experiments were conducted using four NVIDIA RTX A6000 GPUs, each equipped with 48GB of RAM. The total computation time amounted to approximately 2500 GPU hours. The breakdown of GPU hours for different models is as follows:

\noindent Roberta-base experiments: 300 GPU hours

\noindent Roberta-large experiments: 600 GPU hours

\noindent Llama-3-8B experiments: 1600 GPU hours



\subsection{Impact of the Annotators' Disagreement on Performance}
\begin{figure*}[ht]
  \centering
     \begin{subfigure}{\columnwidth}
        \includegraphics[width=\columnwidth]{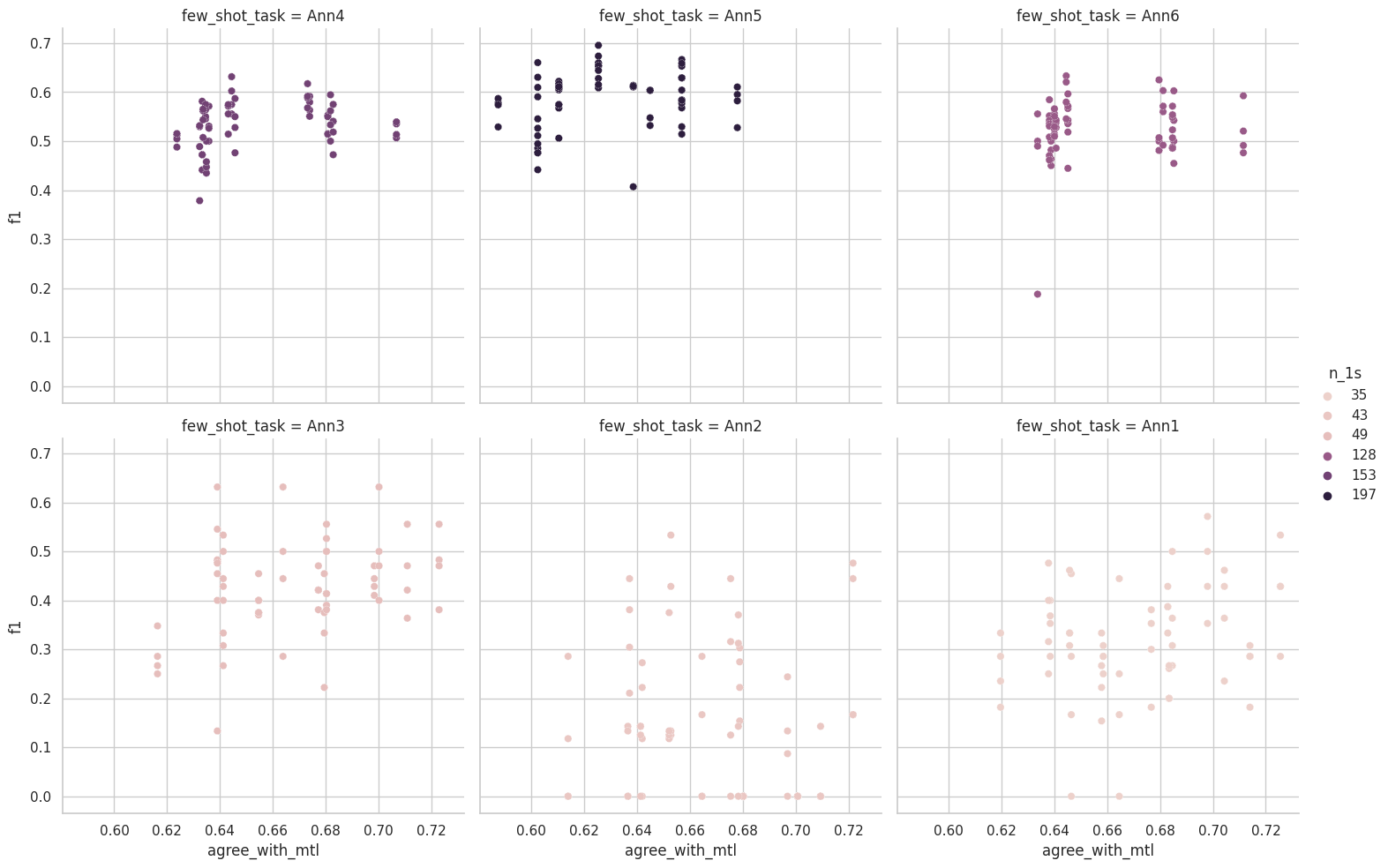}
    \caption{BREXIT}
    \end{subfigure}
   \hfill
       \begin{subfigure}{\columnwidth}
        \includegraphics[width=\columnwidth]{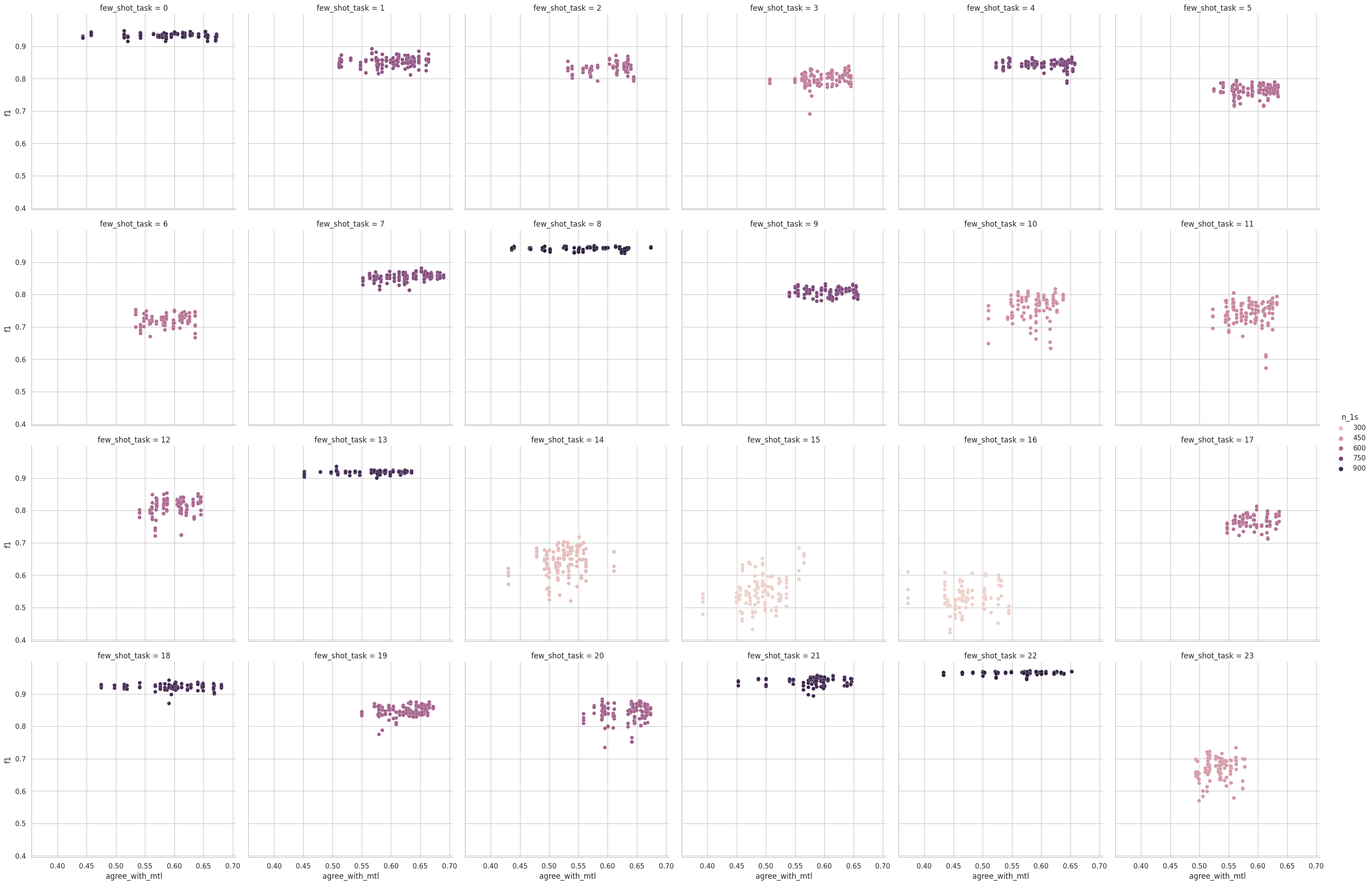}
        \caption{MFSC}
    \end{subfigure}
    \caption{Each plot demonstrates the effect of a single annotator's agreement with the initial set of annotators used for MTL training ($A_{mtl}$), on its $F_1$ score performance, when adopted as a few-shot task. The x-axis represents the agreement measure, and the y-axis represents the $F_1$ score. The darker color of the scatter plot corresponds to a higher number of positive labels provided by the respective annotator.}
    \label{fig:fs-dis-f1}
\end{figure*}

In \autoref{fig:fs-dis-f1} we demonstrate the impact of agreement (as a measure of similarity) between the first and second-stage annotators ($\mathcal{A}_{mtl}$ and $\mathcal{A}_{fs}$) on the performance of the model for the second stage annotators. Importantly, we do not observe performance degradation as the agreement between the two sets decreases.

\section{Mathematical Symbols}
\autoref{tab:math} provides a directory of mathematical symbols used in our paper, along with their respective meanings, to facilitate ease of understanding for the reader.
\setlength{\tabcolsep}{0.7pt}
\begin{table}[t!]
\small
    \centering
    \begin{tabular}{|c|c|}  \hline
          Symbol & Meaning\\ \hline
          $\mathcal{A}_{fs}$& Annotators in MTL model\\  
          $\mathcal{A}_{mtl}$& Annotators adopted as few shot task\\ \hline
          $\mathcal{S}_{mv}$& Sampling based on majority vote\\ 
          $\mathcal{S}_{bal}$& Sampling based on balanced samples across classes\\ 
  $\mathcal{S}_{dis}$&Sampling based on high disagreement of annotaions\\ 
          $\mathcal{S}_{rand}$& Random sampling\\ \hline 
  $B$ & Budget\\ \hline 
  $D$ & All annotations for a dataset\\ \hline
  $F^{fs}_1$ & Avg. $F_1$ scores of the few-shot model for $\mathcal{A}_{fs}$ \\ 
 $F^{mtl}_1$& Avg. $F_1$ scores of the multi-task model for $\mathcal{A}_{mtl}$\\ \hline
    \end{tabular}
    \caption{Mathematical notations used throughout the paper with their explanations}
    \label{tab:math}
\end{table}




\end{document}